\documentclass{article}

\usepackage{xcolor}
\usepackage{arxiv}
\usepackage{CJKutf8}
\usepackage[utf8]{inputenc} 
\usepackage[T1]{fontenc}    
\usepackage{hyperref}       
\usepackage{url}            
\usepackage{booktabs}       
\usepackage{amsfonts}       
\usepackage{nicefrac}       
\usepackage{microtype}      
\usepackage{lipsum}
\usepackage{multirow}
\usepackage{subfigure}
\usepackage{graphicx}
\usepackage{footnote}
\usepackage{amsmath,amsfonts,amssymb}
\usepackage{threeparttable}
\usepackage{CJK}
\makesavenoteenv{tabular}
\makesavenoteenv{table}

\usepackage{subfiles} 
\usepackage{makecell}

\usepackage{array}

\newcommand{\MODEL}{PanGu-$\alpha$}

\usepackage[normalem]{ulem}

\title{\MODEL: Large-scale Autoregressive Pretrained Chinese Language Models with Auto-parallel Computation}
\author{Wei Zeng\thanks{Equal Contribution} \and \textbf{Xiaozhe Ren}$^{*}$ \and \textbf{Teng Su}$^{*}$ \and \textbf{Hui Wang}$^{*}$ \\ \and \textbf{Yi Liao~~~~~Zhiwei Wang~~~~~Xin Jiang~~~~~Zhenzhang Yang} \\ \and \textbf{Kaisheng Wang} \\ \and  \textbf{Xiaoda Zhang}\\ \and \textbf{Chen Li}\\ \and \textbf{Ziyan Gong}\\ \and \textbf{Yifan Yao}\\ \and \textbf{Xinjing Huang} \\ \and \textbf{Jun Wang}\\ \and \textbf{Jianfeng Yu}\\ \and \textbf{Qi Guo}\\ \and \textbf{Yue Yu}\\ \and \textbf{Yan Zhang}\\ \and \textbf{Jin Wang}\\ \and \textbf{Hengtao Tao}\\ \and \textbf{Dasen Yan}\\ \and \textbf{Zexuan Yi}\\ \and \textbf{Fang Peng}\\ \and \textbf{Fangqing Jiang}\\ \and \textbf{Han Zhang}\\ \and \textbf{Lingfeng Deng}\\ \and \textbf{Yehong Zhang}\\ \and \textbf{Zhe Lin}\\ \and \textbf{Chao Zhang~~~~~~Shaojie Zhang~~~~~~Mingyue Guo~~~~~~Shanzhi Gu~~~~~~Gaojun Fan~~~~~~Yaowei Wang} \\ \and \textbf{Xuefeng Jin~~~~~~Qun Liu~~~~~~Yonghong Tian}\\ \\ \\
\textsc{\large \MODEL\ Team}\\
}

\begin{document}
\date{}
\maketitle

\begin{abstract}
Large-scale Pretrained Language Models (PLMs) have become the new paradigm for Natural Language Processing (NLP). PLMs with hundreds of billions parameters such as GPT-3~\cite{brown2020GPT3} have demonstrated strong performances on natural language understanding and generation with \textit{few-shot in-context} learning. In this work, we present our practice on training large-scale autoregressive language models named \MODEL, with up to 200 billion parameters. \MODEL\ is developed under the MindSpore\footnote{\url{https://www.mindspore.cn/en}} and trained on a cluster of 2048 Ascend 910 AI processors\footnote{\url{https://e.huawei.com/en/products/servers/ascend}}. 
The training parallelism strategy is implemented based on MindSpore Auto-parallel,  which composes five parallelism dimensions to scale the training task to 2048 processors efficiently, including data parallelism, op-level model parallelism, pipeline model parallelism, optimizer model parallelism and rematerialization. To enhance the generalization ability of \MODEL, we collect 1.1TB high-quality Chinese data from a wide range of domains to pretrain the model. We empirically test the generation ability of \MODEL\ in various scenarios including text summarization, question answering, dialogue generation, etc. Moreover, we investigate the effect of model scales on the few-shot performances across a broad range of Chinese NLP tasks. The experimental results demonstrate the superior capabilities of  \MODEL\ in performing various tasks under few-shot or zero-shot settings.
\end{abstract}


\section{Introduction}\label{sec:introduction}

Pre-trained Language Models (PLMs) \cite[etc.]{brown2020GPT3,devlin2019bert,CPM2020,yang2019xlnet,liu2019roberta,raffel2020,sun2019ernie,zhang2019ernie,wei2019nezha} have gained great success in the Natural Language Processing (NLP). By learning contextual representation of text from large-scale corpora in a self-supervised manner, PLMs can achieve state-of-the-art performances on a wide range of Natural Language Understanding (NLU) and Natural Language Generation (NLG) tasks.

Radford et. al.~\cite{radford2018improving} demonstrates a significant gains on a variety of NLP tasks via Generative Pre-trained Transformer (GPT), which is an autoregressive language model first pretrained on unsupervised text data and then finetuned for each supervised task. Devlin et.al.~\cite{devlin2019bert} proposes BERT, a bidirectional Transformer with the masked language model (MLM) pretraining objective, which obtains new state-of-the-art performances on the GLUE benchmark of NLU tasks. After them, there have been an increasing number of research work on developing the pretraining techniques and continuously improving the performance of downstream NLP tasks. Among all the techniques, researchers find that the performance of PLMs can be steadily improved simply by enlarging the amount of the training data as well as the capacity of the model. For instance, RoBERTa~\cite{liu2019roberta} shows that BERT can be substantially improved by training the model longer with more data. GPT-2~\cite{radford2019language} as the successor of GPT, which shares the same architecture but contains 1.5 billion parameters and is trained with 40GB text, can perform reasonably well on multiple tasks in the zero-shot setting. The T5 model~\cite{raffel2020} with 11 billion parameters trained on the 745GB C4 data, keeps pushing the performance of both NLU and NLG tasks.

Recently, the OpenAI team announced its lasted version of the GPT-series models: GPT-3~\cite{brown2020GPT3}. The largest GPT-3 model contains 175 billion parameters and is trained using 570GB of text data. Besides its strong capability in generating high-quality text, GPT-3 is especially effective in solving a wide range of tasks without task-specific finetuning in the few-shot, or even zero-shot settings. Moreover, on many of the tasks the performance improves steadily as the size of the GPT model grows, and sometimes even reaches the level of the prior state-of-the-art finetuning approaches. From applications perspective, GPT-3 is revolutionary, as it relieves the need for labelling many examples and retraining model for every new task, which hinders the applicability of NLP models in real-world applications.

However, GPT-3 is now only available for limited access via OpenAI API, and it is primarily trained with English data. 
To promote the public research of Chinese PLMs, we propose training a very large-scale Chinese PLM named \MODEL\, with number of parameters up to 200 billion. To the best of our knowledge, this is the largest Chinese PLM up to the publication of this technical report.

The difficulty in training a PLM rises as the scale of the model grows beyond the level of 10 billion. The main challenges lie in three aspects:

\begin{itemize}
\item \textbf{Model Design}. There have been a couple of architectures of PLMs besides GPT and BERT. However, not all the PLMs can be smoothly scaled to hundreds of billions of parameters. For examples, some models may have problem of slow convergence or even divergence during training as the model size increases. Inspired by GPT-3 and our preliminary experiments, we choose the Transformer-based autoregressive language model as the base architecture. Besides, we develop an additional \textit{query layer} on top of the Transformer layers to induce the expected output of the model during pretraining. Our experiments demonstrate that the structure of \MODEL\ can scale up to 200 billion parameters. 

\item \textbf{Training Corpora}. Training data is essential in building a strong and generalisable pretrained model. On one hand, the amount of the data should be sufficient to feed a large PLM. On the other hand, the data should be of high quality and diversity to ensure the generality of the PLM. To build Chinese corpus with comprehensive coverage, we collect a large amount of data from a wide range of resources, including Common Crawl, e-Books, encyclopedias, news, and so on. Based on them, we conduct multiple processes of data filtering and cleaning to make sure the processed data are of high quality and reliability.

\item \textbf{Distributed Training.} The memory requirement of training \MODEL\ with 200 billion parameters is much beyond the memory capacities of modern AI processors. It is difficult to acquire large end-to-end throughput while keeping high resource utilization on a cluster of processors. The problem becomes more challenging when considering the topology of hardware. We combine five-dimensional parallel functionalities with a carefully designed parallelization strategy and apply them to the largest \MODEL, which is efficiently
trained on a cluster of 2048 Ascend 910 AI processors~\cite{ascend-davinci} and powered by CANN\footnote{\url{https://www.hiascend.com/en/software/cann}}. 
\end{itemize}

We train three \MODEL\ models on a high-quality 1.1TB Chinese text corpus with increasing magnitude of parameter sizes, which are \MODEL\ 2.6B, \MODEL\ 13B, and \MODEL\ 200B, respectively. We first evaluate the models on language modeling tasks, showing that the perplexity can be decreased with the increase of model capacity and the amount of data and computation. Then we investigate the text generation ability of \MODEL\ in various scenarios such as dialogue generation, summarization, question answering, etc. We demonstrate a few generated samples for different applications in the experiment section. Furthermore, we evaluate the task-agnostic few-shot performances of \MODEL\ 2.6B and 13B on a wide range of NLP tasks, including cloze tasks, reading comprehension, closed-book QA, Winograd style tasks, commonsense reasoning, natural language inference, and text classification. The experimental results demonstrate that with the growing model capacity, the performance on various tasks can generally improve.

We are currently seeking a proper way to let both non-profit research institutes and commercial companies to get access to our pretrained \MODEL\ models, either by releasing the code and model or via APIs. We are also assessing the possibility of releasing all or part of our pretraining data, within the constraints of the law and legality.

To facilitate the community to pretrain a large-scale language model by their own, the parallel computing functionalities are open-sourced in the Auto-parallel module of MindSpore\footnote{\url{https://gitee.com/mindspore/mindspore}}, a deep learning training/inference framework that could be used for mobile, edge and cloud scenarios. Besides the basic parallel functionalities, Auto-parallel is easy enough to use by freeing developers from parallel model training with minimal (or zero) code modifications from the standalone version, \textit{as if} the model is trained on a single device.  


The reminder of this technical report is organized as follow. Section~\ref{sec:model} describe the architecture of our \MODEL\ models. In section~\ref{sec:dataset}, we detail our methods to construct a 1.1TB high-quality training corpus from 80TB raw data collected from various sources. Section~\ref{sec:system} addresses the parallelization paradigm of model training and scheduling strategy on a cluster of Ascend processors. Section~\ref{sec:experiments} presents the experimental results of \MODEL\ models on various tasks.

\section{Model}\label{sec:model}


\subsection{Overview}
\MODEL\ is a large-scale autoregressive language model (ALM) pretrained on a large corpus of text, mostly in Chinese language. It models the generative process of all the tokens in the corpus, where the generation of a token depends on its previous tokens in a sequence. Assuming that a sequence $X=\{x_1,x_2,...,x_N\}$ is composed of $N$ tokens, the training objective can be formulated as maximization of the log-likelihood:
\begin{equation} \label{ALM}
\mathcal{L} = \sum_{n=1}^{N}\log p(x_n|x_1, ..., x_{n-1};\theta),
\end{equation}
\noindent where $p(x_n|x_1, ..., x_{n-1};\theta)$ is the probability of observing the $n$-th token $x_n$ given the previous context $x_{1:n-1}$, and $\theta$ denotes the model parameters. 

\begin{figure*}[]
	\centering
	\includegraphics[trim=0 2cm 0 2cm, width=0.9\textwidth]{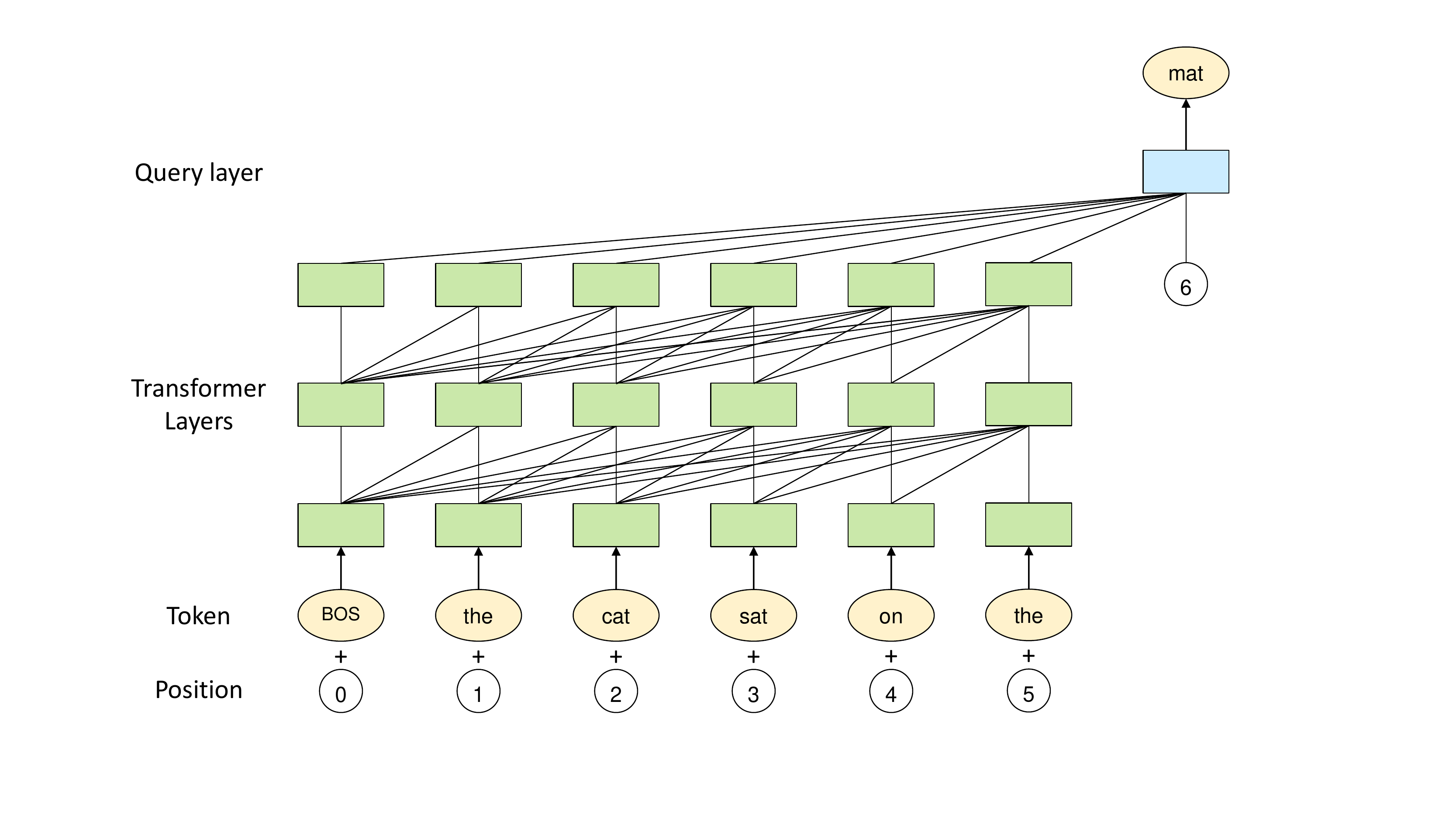}
	\caption{The architecture of \MODEL. The model is based on a uni-directional Transformer decoder. A query layer is stacked on top of Transformer layers with the position embedding as the query in the attention mechanism to generate the token at the next position.}
	\label{fig:overview}
\end{figure*}

The architecture of \MODEL\ is based on Transformer~\cite{vaswani2017attention}, which has been extensively used as the backbone of a variety of pretrained language models such as BERT~\cite{devlin2019bert} and GPT~\cite{radford2018improving,radford2019language,brown2020GPT3}. Different from them, we develop an additional \textit{query layer} on top of Transformer layers to predict the next token. The diagram of the model is shown in Figure~\ref{fig:overview}. We elaborate each part as follow.

\subsection{Model Structure}
\subsubsection{Transformer Layers}
A standard transformer layer includes two sub-layers: multi-head attention (MHA) and fully connected feed-forward network (FFN).

\textbf{Multi-head Attention:} A self-attention network in the $l$-th Transformer layer is parameterized by four projection matrices: $W^k_h, W^q_h, W^v_h, W^m_h \in \mathbb{R}^{d \times d/N_{h}}$, where $d$ is the hidden dimension, $h$ is the index of head, and $N_h$ is the number of heads. Given the output $H_{l-1} \in \mathbb{R}^{N \times d}$ from the precedent layer, three major components, i.e., query $Q_h=H_{l-1}W^q_h$, key $K_h=H_{l-1}W^k_h$, and value $V_h=H_{l-1}W^v_h$ are produced. The attention function is computed as:
\begin{equation}
\begin{split}
    A_h & = Q_hK_h^\top = H_{l-1}W_h^qW_h^{k\top}H_{l-1}^{\top},\\
    \text{Attention}_h(H_{l-1}) &= \text{Softmax}(\frac{A_h}{\sqrt{d}})V_h = \text{Softmax}(\frac{A_h}{\sqrt{d}})H_{l-1}W_h^v.
\end{split}
\end{equation}
With multiple attention heads, the output becomes:
\begin{equation}
\begin{split}
    \text{MHA}(H_{l-1}) &= \sum_{h=1}^{N_h} \text{Attention}_h(H_{l-1})W^m_h,\\
    H_{l}^{\text{MHA}} &= H_{l-1}+\text{MHA}(\text{LayerNorm}(H_{l-1})).
\end{split}
\end{equation}

\textbf{Feed-forward Network:} The FFN layer is composed of two linear layers, parameterized by $W^1 \in \mathbb{R}^{d \times d_{ff}}$, $b^1 \in \mathbb{R}^{d_{ff}}$, $W^2 \in \mathbb{R}^{d_{ff} \times d}$, $b^2 \in \mathbb{R}^{d}$, where $d_{ff}$ is the dimension of the inner-layer. Fed with the output of MHA layer as input, the output of FFN layer is then computed as:
\begin{equation}
\begin{split}
\text{FFN}(H_l^{\text{MHA}}) &= \text{GeLU}(H_l^{\text{MHA}}W^1 + b^1)W^2 + b^2,\\
H_{l} &= H_{l}^{\text{MHA}}+\text{FFN}(\text{LayerNorm}(H_{l}^{\text{MHA}})).
\end{split}
\end{equation}

For both MHA and FFN, we take the \textit{pre-layer normalization} scheme, which can make the training of Transformer model easier and faster~\cite{xiong2020layer}.

\subsubsection{Query Layer}
We design the query layer on top of the stacked Transformer layers, which aims to explicitly induce the expected output. In the pretraining stage of the autoregressive model, it comes to the prediction of the next token. The structure of the query layer resembles the transformer layer, except that an additional embedding $p_n \in \mathbb{R}^d$ indicating the next position is used as the query vector in the attention mechanism. Specifically, assuming $H_L$ is the output of the uppermost transformer layer, the attention vector in the query layer is computed as:
\begin{equation}
    a_h = p_nW_h^qW_h^{k\top}H_{L}^{\top}.
\end{equation}
The subsequent computation of $\text{MHA}$ and $\text{FFN}$ remains the same as the original Transformer. We denote the final output as $o_n$. The negative log-likelihood of next token becomes:
\begin{equation}
\begin{split}
\text{CrossEntropy}(x_n, \text{Softmax}(o_nW^o+b^o)),
\end{split}
\end{equation}
where $x_n$ denotes the true token and $W^o, b^o$ is the additional task-dependent parameters.

\begin{table*}
\centering
  \caption{Model sizes and hyperparameters of \MODEL\ models.}
  \label{tab:config-pangu}
  \begin{tabular}{cccccc}
     \toprule
    Model & \#Parameters & \#Layers ($L$) & Hidden size ($d$) & FFN size ($d_{ff}$) & \#Heads ($N_h$) \\
     \midrule
    \MODEL\ 2.6B & 2.6B & 32 & 2560 & 10240 & 40  \\
    \MODEL\ 13B  & 13.1B & 40 & 5120 & 20480 & 40 \\
    \MODEL\ 200B & 207.0B & 64 & 16384 & 65536 & 128 \\
  \bottomrule
\end{tabular}
\end{table*}

\subsubsection{Model Configurations}
To evaluate the scaling ability of the \MODEL\ model, we train three models with increasing magnitude of parameter sizes, that is, \MODEL\ 2.6B, \MODEL\ 13B, and \MODEL\ 200B. Table~\ref{tab:config-pangu} shows the detailed configurations of the three models, including the number of total parameters, the hidden dimension for the tokens, the inner dimension of the feed-forward layer, and the number of attention heads.

\section{Dataset}\label{sec:dataset}


A large-scale Chinese text corpus of high quality is crucial for the pretraining of our \MODEL\ models, especially the one with 200B parameters.
Existing large-scale text corpora for pretraining super large language models are mainly English.
For example, the GPT-3 \cite{brown2020GPT3} is trained using a dataset which contains 570GB filtered texts from Common Crawl with $92.6\%$ of the words are English. The \emph{Colossal Clean Crawled Corpus} (C4) for training T5 consists of about 750GB clean English texts scraped from the web \cite{raffel2020}.
To the best of our knowledge, there are three Chinese text corpora that are above 100GB: (a) CLUECorpus2020 (100GB), which is retrieved from the Common Crawl dataset \cite{xu2020cluecorpus2020}; (b) the Chinese multi-modal pretraining data, released by \cite{lin2021m6} which contains 300GB texts; and (c) WuDaoCorpus\footnote{\url{https://data.baai.ac.cn/data-set-details/0c8dc71dd06ae75a10ca422fb49b0751}}, which opens about 300GB text data to only specific partners so far.
However, all the above datasets are still not enough to train the super large-scale models up to 200B parameters compared to the data size used in existing English pretrained models.

Even though the raw web datasets such as SogouT\footnote{\url{https://www.sogou.com/labs/resource/t.php}} and Common Crawl\footnote{\url{https://commoncrawl.org/the-data/}} contain massive amount of Chinese texts, the construction of our desired dataset is still challenging due to the highly varying quality of the raw web data, the huge amount of storage and computation to preprocess the data, and the lack of well-defined metrics to evaluate the quality of the data.

To tackle the aforementioned issues, we construct a 1.1TB high-quality Chinese text corpus by cleaning and filtering enormous raw data from multiple sources. A \emph{big data management platform} is built to accelerate the massive data analysis and processing. Both manual and model-based evaluation measures are used to guide the data preprocessing and training data selection, as detailed in the following sections.

\subsection{Dataset Construction}

\begin{figure}[t]
\centering
\includegraphics[scale=0.5]{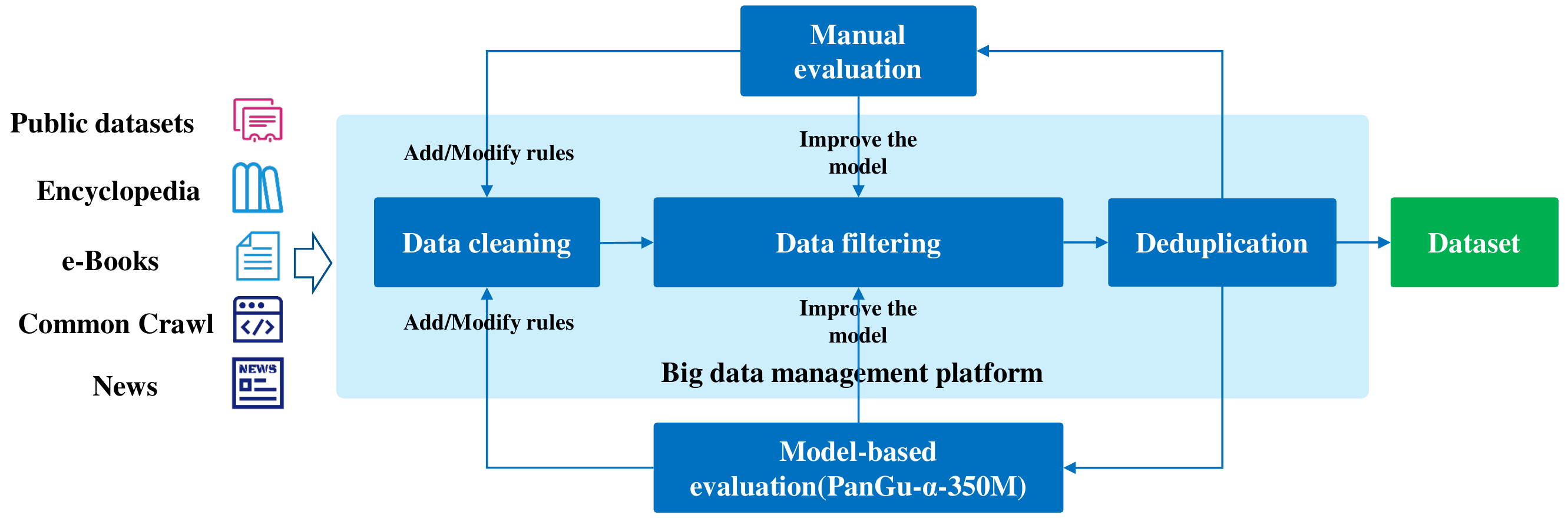}
\caption{The data sources and the process of constructing pretraining data for \MODEL.}
\label{fig:data_process_pipeline}
\end{figure}

To construct a large-scale high-quality Chinese corpus, we collect nearly 80TB raw data from the public datasets (e.g., BaiDuQA, CAIL2018, Sogou-CA, etc.), web pages data from Common Crawl, encyclopedia, news and e-books. As shown in Figure~\ref{fig:data_process_pipeline}, our data construction process includes three steps: rule-based data cleaning, model-based data filtering and text deduplication. To improve the quality of the training dataset, the first two steps (i.e., cleaning and filtering) are iteratively enhanced via manual and model-based data quality evaluations.
The data construction process is done on a \emph{big data management platform} built based on the open source Spark/Hadoop framework using 8 high-performance computing nodes\footnote{4 computing nodes with 28TB storage + 2 CUPs (24 cores) + 1.5TB Memory and 4 computing nodes with 7.3TB storage + 2 CPUs (64 cores) + 1TB memory.}.
With the distributed processing capability and the tools of our platform, the efficiency of the data analysis and processing is significantly improved (see Table~\ref{tab:time-efficiency} for the processing time). 
Next, we introduce the details of each step in the dataset construction process.

\begin{table}[t]
\renewcommand\arraystretch{1.3}
\caption{Processing time for each step in the dataset construction.}
\centering
\begin{tabular}{c|c|c}
 \Xhline{1pt}
  & Data size 
 & Our platform \\ \hline
 Cleaning & 20TB & 70+ hours \\ \hline
 Filtering & 800GB & 10+ hours \\ \hline
Fuzzy deduplication & 500GB  & 3.5 hours \\ 

\Xhline{1pt}
\end{tabular}
\end{table}\label{tab:time-efficiency}

\subsubsection{Cleaning and Filtering}

Among the five data sources as shown in Fig~\ref{fig:data_process_pipeline}, the Common Crawl data contributes the most amount to our corpus but unfortunately contains a significant amount of low-quality web pages. To improve the data quality,
we first adopt the following rule-based text cleaning strategies over the raw web pages from Common Crawl:
\begin{itemize}
    \item Remove the document which contains less than 60\% Chinese characters, or less than 150 characters, or only the title of a webpage;
    \item Remove the special symbols and duplicated paragraphs in each document;
    \item Identify advertisements based on keywords and remove documents which contain advertisements;
    \item Convert all traditional Chinese text to simplified Chinese;
    \item Identify the navigation bar of the web page and remove it.
\end{itemize}


Then, three filters are applied to the preprocessed documents to further remove the harmful, advertising and low-quality documents.
\begin{itemize}

    \item \emph{Sensitive word filtering}: The original documents of Common Crawl include a lot of harmful or sensitive website contents which would mislead our generative model. Thus, we manually collect 724 sensitive words and remove documents containing more than three of the sensitive words.
    
    \item \emph{Model-based spam filtering}: To further remove the advertisements and spams, we train a spam classification model using fastText\footnote{\url{https://fasttext.cc/}} on a manually labeled dataset. The negative training examples are 10K junk documents manually selected from the Common Crawl dataset, and the positive examples are sampled from the high-quality Chinese text corpus. We remove the documents that are classified as spams.
    
    \item \emph{Low-quality document filtering}: Following the practice in GPT-3, we train a classifier to score the quality of each document and eliminate the documents with scores below a threshold (see Appendix~A of \cite{brown2020GPT3} for details).
\end{itemize}

\subsubsection{Text Deduplication}

Although we have removed duplicated paragraphs in each document in the previous step, there are still documents with highly overlapped content across different data sources. Therefore, we carry out fuzzy data deduplication over the documents across all our data sources.

Due to the super large scale of the whole dataset, the conventional MinHashLSH algorithm in Spark incurs more than 8 hours to duplicate less than 200MB data, which is too slow to meet our efficiency requirement.
To accelerate the deduplication process, we design a \emph{distributed large-scale text data duplication detection and deduplication algorithm} by exploiting the computing framework of our big data management platform. 
%
The proposed algorithm takes only 3.5~hours to complete the deduplication process for 500GB documents.

\subsubsection{Data Quality Evaluation}\label{sec:data-evaluation}

Give above preprocessing steps, one key question is how the cleaning rules and the filtering thresholds are decided. In this work, we evaluate the data quality after each round of preprocessing and update the cleaning rules and the filtering models according to the evaluation results. Both manual and model-based evaluations are considered. The manual evaluation is conducted over randomly sampled texts from the perspectives of sentence smoothness and the amount of low-quality contents (e.g., advertisements, repeated short sentences, spams, etc.). 
However, the manual evaluation can only cover a very small proportion of the whole dataset. To improve the accuracy of the data evaluation, we train the \MODEL\ 350M model using 30GB data sampled from the preprocessed dataset and evaluate the data quality using the PPL on a high-quality development dataset. The preprocessed dataset that achieves lower PPL is considered to have higher quality and its corresponding cleaning rules and filtering models are considered to be better.

\subsection{Training Data Selection}

\begin{savenotes}
    \begin{table}[t]
    \renewcommand\arraystretch{1.3}
    \caption{Data composition of the 1.1TB Chinese text corpus.}
    \centering
        \begin{tabular}{c|c|c|c}
        \Xhline{1pt}
         & Size (GB) & Data source & Processing steps \\ \hline
        Public datasets &	27.9 & \multicolumn{1}{m{6cm}|}{15 public datasets including DuReader, BaiDuQA, CAIL2018, Sogou-CA, etc.} & \multicolumn{1}{m{4cm}}{Format conversion\footnote{We remove the labels in all the labeled datasets such that the model is trained for few-shot learning instead of multi-task learning.} and text deduplication} \\ \hline
        Encyclopedia &	22 & \multicolumn{1}{m{6cm}|}{Baidu Baike, Sogou Baike, etc.} &	Text deduplication\\ \hline
        e-Books &	299 & \multicolumn{1}{m{6cm}|}{e-Books on various topics (e,g., novels, history, poetry, ancient prose, etc.).} &	\multicolumn{1}{m{4cm}}{Sensitive word and model-based spam filtering}\\ \hline
        Common Crawl &	714.9 &	\multicolumn{1}{m{6cm}|}{Web data from January 2018 to December 2020 from Common Crawl.} & All steps \\ \hline
        News &	35.5 &	News data from 1992 to 2011. & Text deduplication\\
        \Xhline{1pt}
        \end{tabular}
    \end{table}\label{tab:data-composition}
\end{savenotes}

\begin{figure}[t]
	\centering
	\begin{tabular}{cc}
		\hspace{-1mm}\includegraphics[scale=0.38]{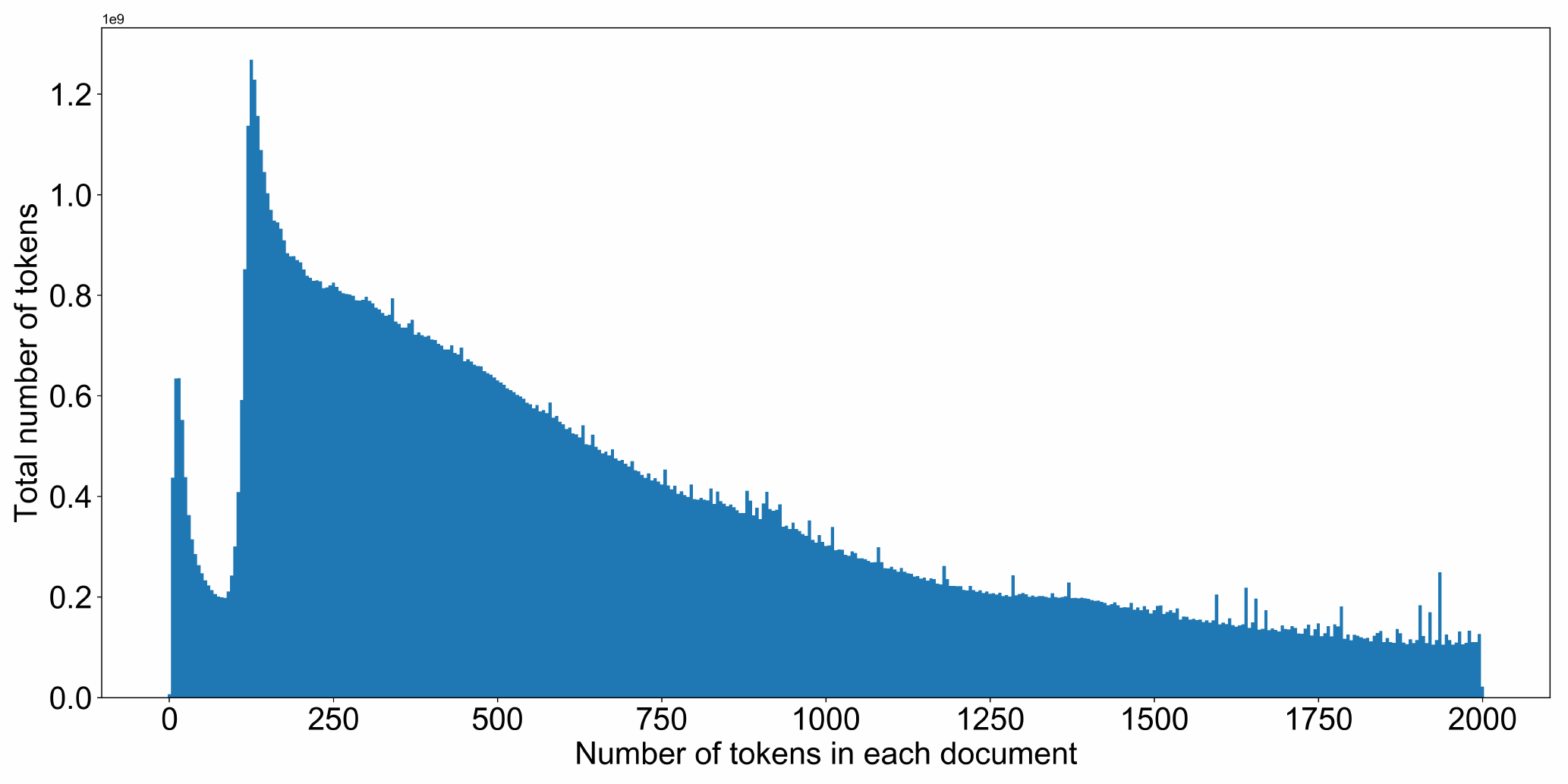} &
		\hspace{-1mm}\includegraphics[scale=0.38]{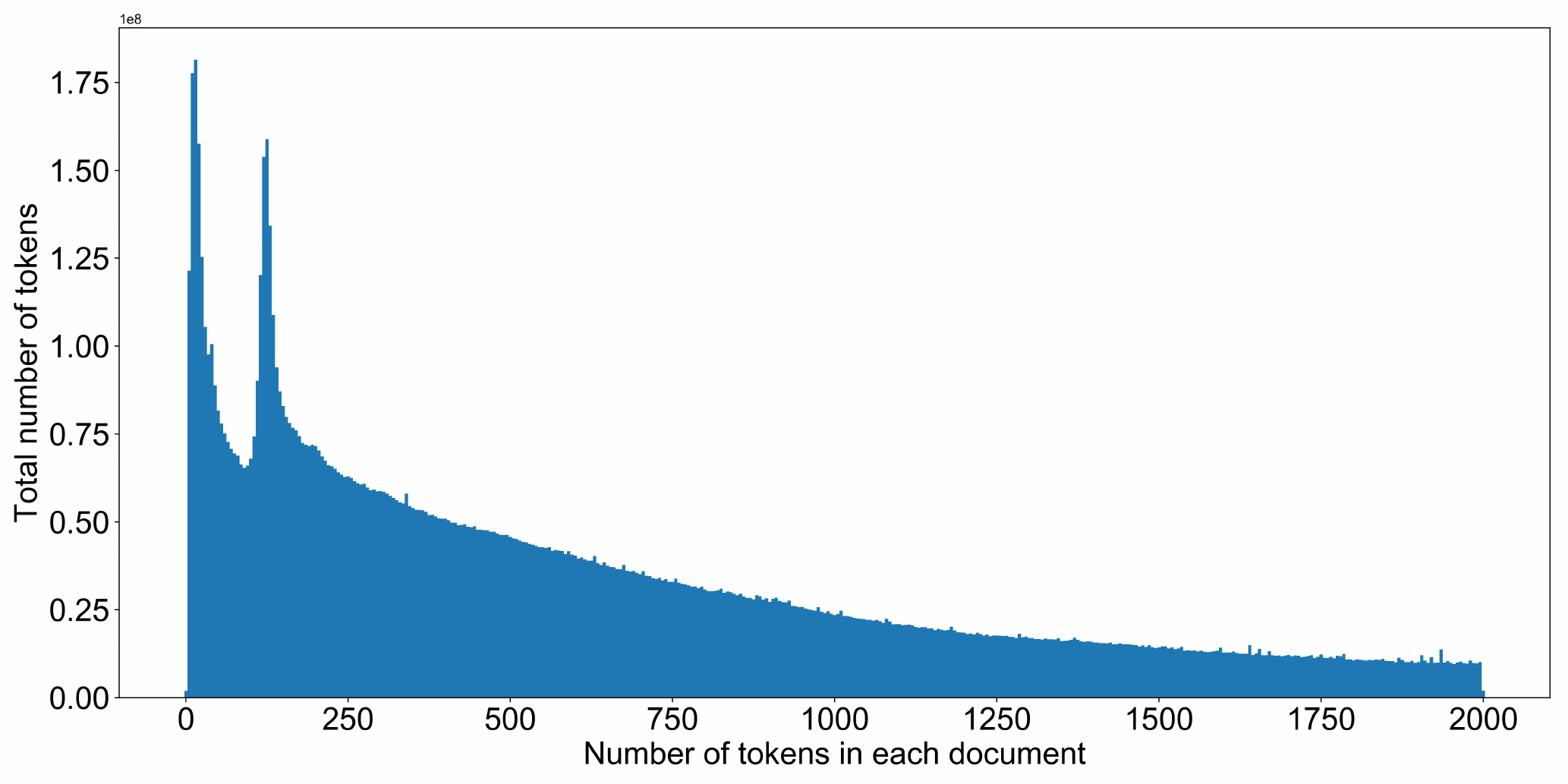} \\
		\hspace{-1mm} (a) 1TB dataset & \hspace{-1mm} (b) 100GB dataset \\
	\end{tabular}
	\caption{The distribution of tokens in (a) 1TB dataset and (b) 100GB dataset. The total number of tokens represents the (number of tokens in each document) $\times$ (number of documents with this token number).} 
	\label{fig:token}
\end{figure}

Using the construction process in Figure~\ref{fig:data_process_pipeline}, a Chinese text corpus with 1.1TB data is built from the five types of data sources. The composition of our corpus and the processing steps adopted to each data source is shown in Table~\ref{tab:data-composition}.
Based on the new corpus, we construct two training datasets with 100GB and 1TB text data for our medium (2.6B and 13B) and large (200B) models, respectively.
As shown in Table~\ref{tab:traindata}, each data source is sampled during training with different proportions according to the quality of the processed dataset evaluated using the method in Section~\ref{sec:data-evaluation}.
The distribution of the number of token in each training dataset is shown in Figure~\ref{fig:token}. The averaged document lengths of the 100GB and 1TB dataset are 239 and 405 tokens, respectively. The 1TB dataset has a larger averaged document length due to the large proportion of Common Crawl dataset.
Note that the length of the text will affect the generation performance of the model. When the averaged number of token for the training samples is small, the model will be biased to generate short texts and be good at processing downstream tasks requiring short texts, and vice versa.

\begin{table}[t]
\renewcommand\arraystretch{1.3}
\caption{Sampling strategy of the corpora in training \MODEL\ models.}
\centering
\newcommand{\tabincell}[2]{\begin{tabular}{@{}#1@{}}#2\end{tabular}}
\begin{tabular}{c|c|c|c|c|c}
\Xhline{1pt}
\multirow{2}{*}{} & \multicolumn{3}{c|}{\MODEL\ 200B} & \multicolumn{2}{c}{\MODEL\ 2.6B\&13B} \\ \cline{2-6} 
 & \tabincell{c}{Quantity \vspace{-1mm}\\(tokens)} &	\tabincell{c}{Weight in \vspace{-1mm}\\ training mix} & \tabincell{c}{Epochs elapsed \vspace{-1mm}\\ when training} & \tabincell{c}{Quantity \vspace{-1mm}\\(tokens)} & \tabincell{c}{Weight in \vspace{-1mm}\\ training mix} \\ \hline
Public datasets  &	25.8B  &	10.23\%  &	3.65 & 7B &	27.99\%\\ \hline
e-Books  &	30.9B  &	12.23\%	  &0.41 & 5.6B & 18\% \\ \hline
Common Crawl  &	176.2B  &	62.81\%  &	0.85 & 2.5B & 10\%\\ \hline
News  &	19.8B  &	7.83\%  &	2.2  & 5.6B & 22\%\\ \hline
Encyclopedia data & 5.8B & 6.9\% & 3 & 5.8B &	23\% \\ \Xhline{1pt}
\end{tabular}
\end{table}\label{tab:traindata}

\section{System}\label{sec:system}

Training \MODEL\ 200B and using it for inference are difficult. The memory requirement for just storing \MODEL\ 200B is around 750 GB. Training such a huge
model consumes several times more memory than just storing the parameters, since the gradients and optimizer states are
also essential for updating the parameters. As a contrast, the memory of modern AI processors (\textit{e.g.}, GPU, Ascend 910 AI processor~\cite{ascend-davinci}) is still around 30-40 GB. Thus, it is inevitable to partition the model to a collection of devices (processors). The problem is challenging in two perspectives. First, multiple basic parallel functionalities should be combined to acquire the end-to-end high performance. Finding the best combination strategy is challenging due to the huge strategy space. Second, parallel training should be easy to use, and the underlying parallel-related code should be removed from the model definition code. We use Auto-parallel in MindSpore to address the problem by \textit{maximizing the ratio of the computation over the communication}. Auto-parallel supports \textit{five-dimensional} parallel functionalities, and employs topology-aware scheduling to map partitioned model slices to the cluster for the end-to-end high performance. Furthermore, Auto-parallel enables the least code modifications from the standalone version for parallel training. 

\subsection{Five-dimensional Parallelisms and Topology-aware Scheduling}
\begin{figure*}[tbp]
	\centering
	\includegraphics[width=0.98\textwidth]{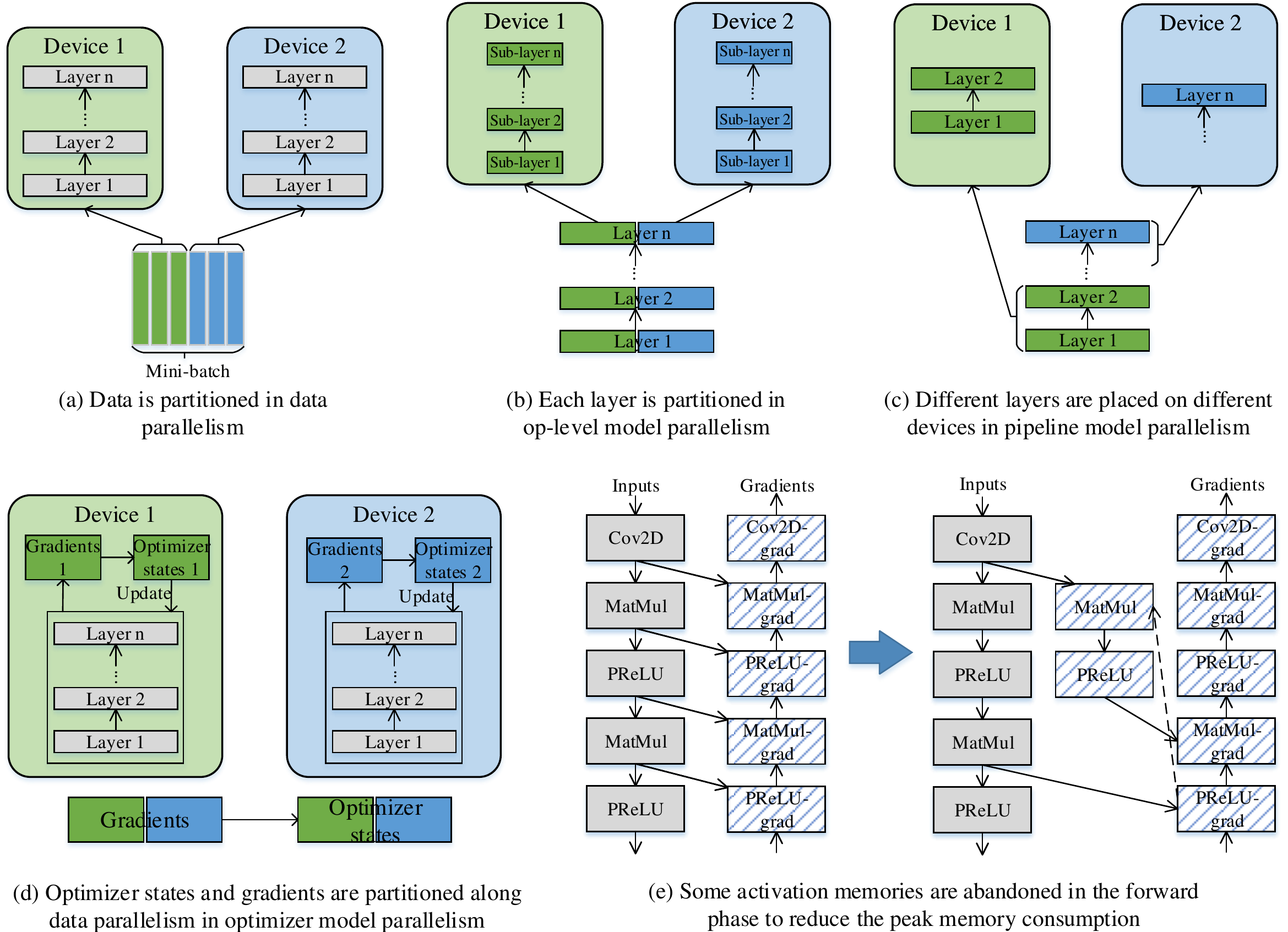}
	\caption{Five parallel functionalities, and how each works to optimize memory and throughput.}
	\label{fig:parallel-compare}
\end{figure*}

The most applied parallelism way is data parallelism, which partitions the training batches across devices, and synchronizes the gradients from different devices before taking an optimizer step, as shown in Figure~\ref{fig:parallel-compare}(a). There are three regimes in model parallelism. One regime is op-level model parallelism~\cite{mesh-tf, pmlr-v80-jia18a, tofu-19, gshard, Megatron, AccPar, flexflow}, which partitions its involved tensors of each operator (layer), as shown in Figure~\ref{fig:parallel-compare}(b). Op-level  model parallelism reduces the memory consumption by slicing the parameters and the activation memory, however, it introduces communications to keep the distributed tensor layouts consistent between successive operators. The second regime is pipeline model parallelism~\cite{Gpipe, PipeDream, DAPPLE, NEURIPS2020_b14680de, HetPipe}, which partitions the total layers to stages, and then places stages to different devices, as shown in Figure~\ref{fig:parallel-compare}(c). The memory benefit comes from that each device holds a subset of total layers of the model, and the communications only happen at the boundaries of stages.
The third regime is optimizer model parallelism~\cite{zero} (Figure~\ref{fig:parallel-compare}(d)), which aims to reduce the redundant optimizer memory and computation consumption resulted from data parallelism.  Some outputs of operators in forward phase reside in memory for a fairly long time, because they are used in the backward phase for gradient calculations. Rematerialization (Figure~\ref{fig:parallel-compare}(e)) abandons these memories to reduce the peak memory consumption in the whole training time, by recomputing the corresponding forward operators.

Each parallelism dimension trades computation (or communication) overheads for memory (or throughput) benefits.
To acquire maximum end-to-end throughput, a balanced composition point should be found along these dimensions. The problem becomes more challenging when considering the heterogeneous bandwidths in a cluster of devices.

\begin{figure*}[tbp]
	\centering
	\includegraphics[width=0.99\textwidth]{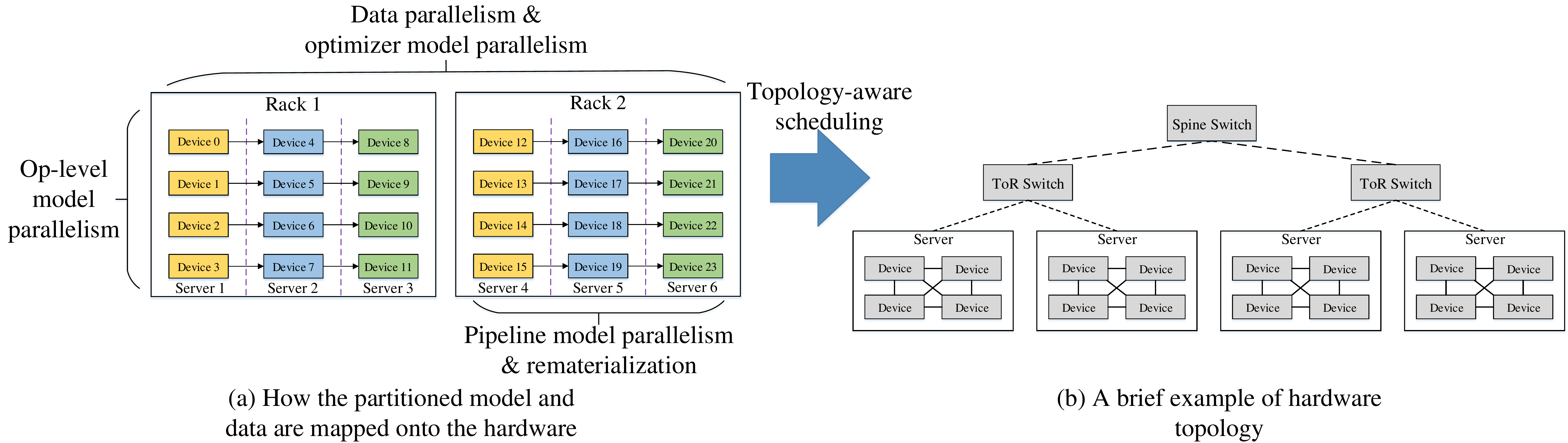}
	\caption{A combined parallelization of the model, and how it is scheduled to the cluster.}
	\label{fig:topology-aware}
\end{figure*}

Figure~\ref{fig:topology-aware}(b) demonstrates a typical organization of a  cluster. Each server includes multiple devices, and the servers in a rack are connected by a ToR (top of rack) switch. Racks are then connected by the Spine switch. The bandwidth between devices in a server is greater than that across servers in a rack, and the latter one is greater than that across racks. Therefore, the model is partitioned across servers in a rack using the pipeline parallelism regime, resulting in that each server holds a stage of the model layers. Then, the stage is split using the op-level parallelism across the devices in each server, in order to utilize the high bandwidths. Each rack owns the whole model, and different racks are data parallel. Deploying data parallelism and optimizer parallelism across racks is due to  that the induced communication operators are not on the critical path of the training iteration, 
which could be fused and overlapped with backward propagation to improve the performance.

\begin{figure*}[tbp]
	\centering
	\includegraphics[width=0.99\textwidth]{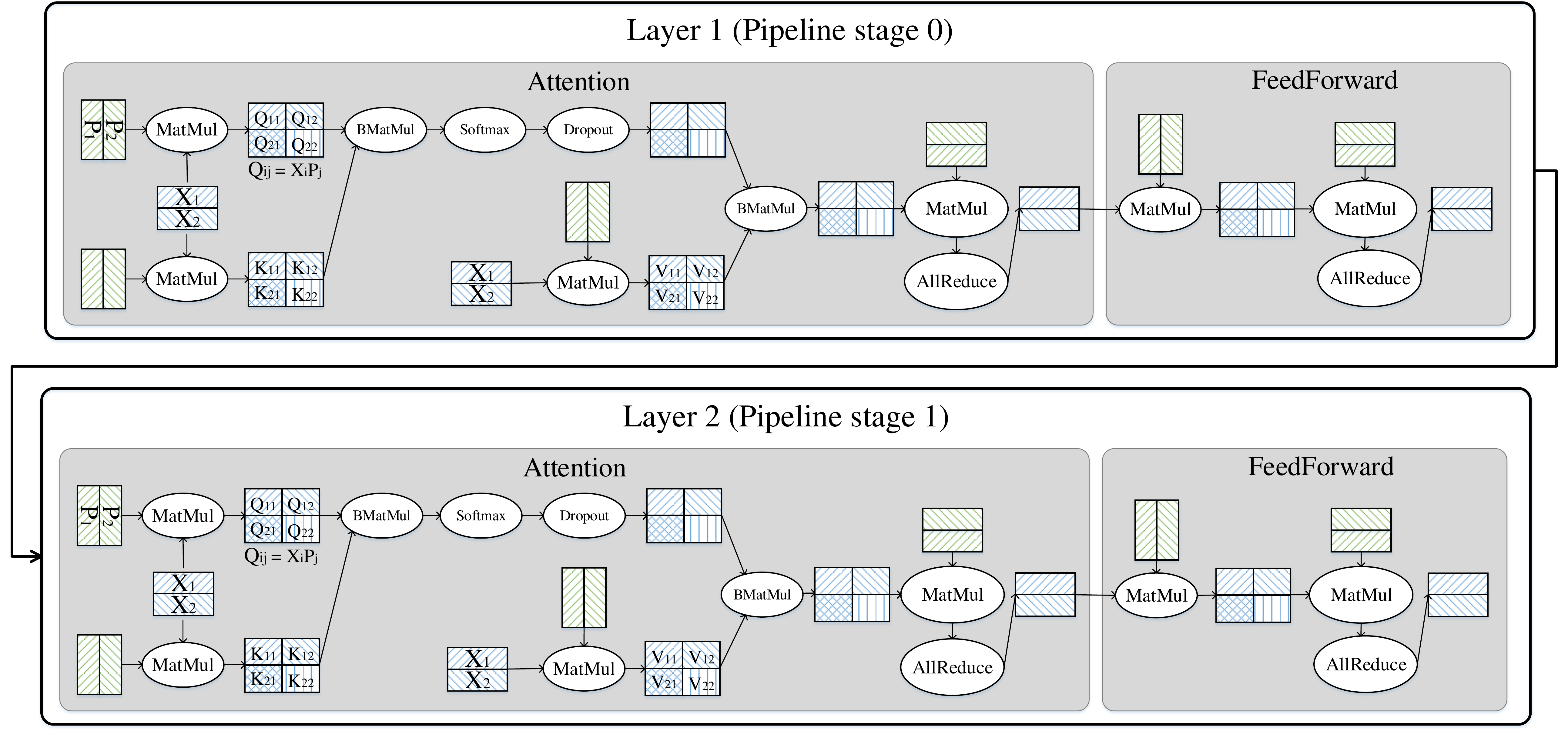}
	\caption{A simplified \MODEL's parallelization strategy. The ellipsoids stand for the operators, blue rectangles represent tensors, and green rectangles represent trainable parameters. Parameters are partitioned along the row  and column dimension respectively, and the input tensor is partitioned along the row dimension. And, two layers are assigned to different pipeline stages.}
	\label{fig:pangu-parallel}
\end{figure*}

Figure~\ref{fig:pangu-parallel} shows how a combined parallelization is applied to the \MODEL\ 200B model. First, 64 layers of the model are partitioned into 16 stages, each stage containing 4 layers. For each layer, involved parameters and tensors are partitioned for each operator. Specifically, the parameters involved in query ($Q$), key ($K$) and value ($V$) operators are partitioned into 8 slices. The input tensor of these three operators is partitioned into 16 slices, and the number of optimizer model parallelism is determined accordingly.\footnote{The `8' is called model parallel number and `16' is called data (and optimizer) parallel number in our system. In the example of  Figure~\ref{fig:pangu-parallel}, the model parallel number and data parallel number are both 2.} Parallelization strategies for other operators in the layer are configured likewise. Rematerialization is configured to perform within each layer, which limits the extra computation overheads. Totally, 
2048 Ascend 910 AI processors are used to train the full \MODEL\ 200B model.

\subsection{Implementation}
The parallel-related functionalities are implemented in the Auto-parallel module of MindSpore. The Auto-parallel decouples machine learning models from complicated underlying parallel implementations, and let researchers focus on the development of new models. Auto-parallel enables parallel training by just adding annotations on the standalone model script. Here, we briefly go through two model parallelism regimes.

\begin{figure*}[tbp]
	\centering

	\includegraphics[width=0.99\textwidth]{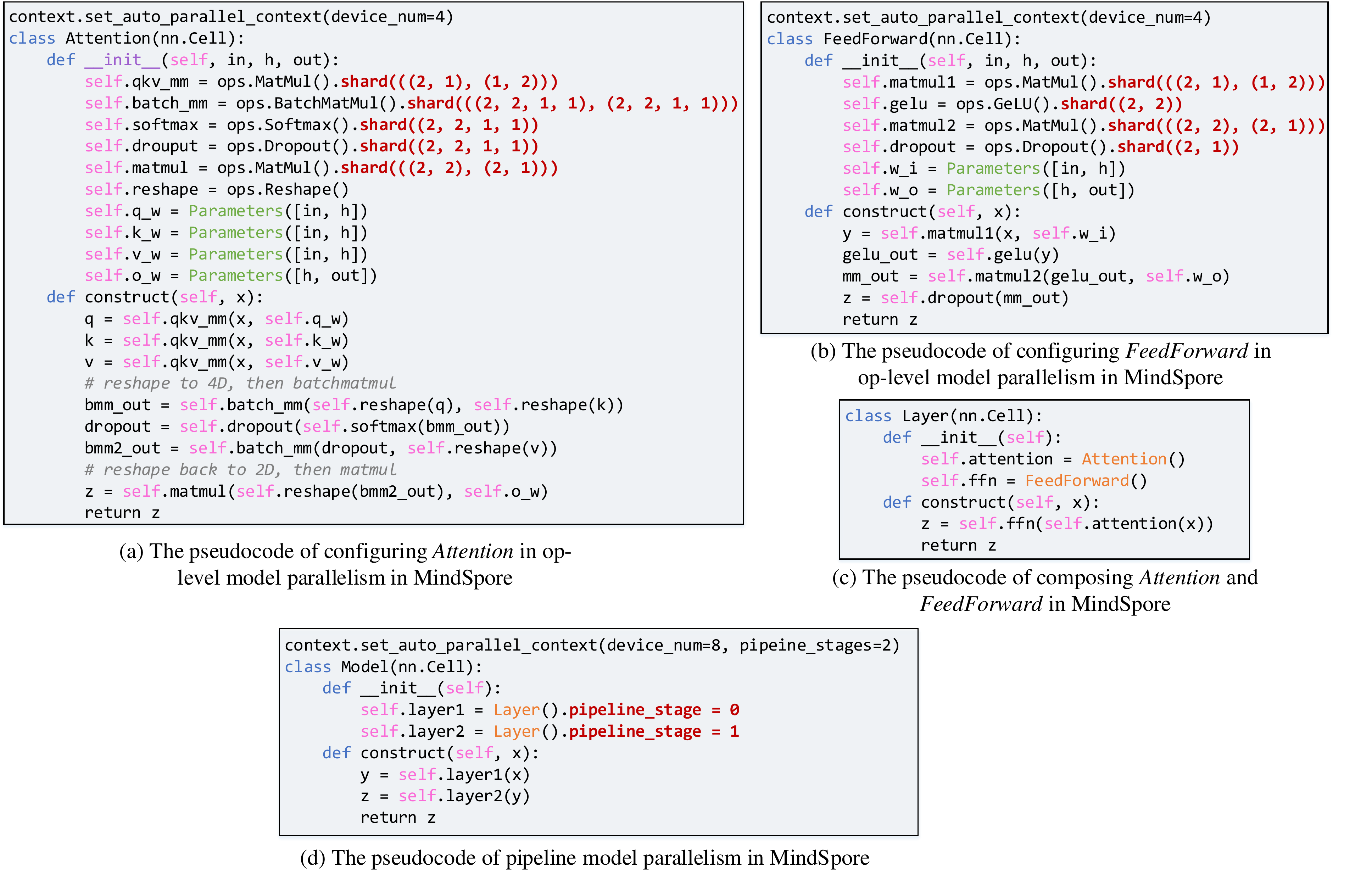}
	\caption{The pseudocode of configuring op-level and pipeline  parallelism in MindSpore. The red bold fonts are keywords to specify parallelization strategies.}
	\label{fig:parallel-usage}
\end{figure*}

Figure~\ref{fig:parallel-usage} shows how to specify the combined parallelization strategy to \MODEL.
Figure~\ref{fig:parallel-usage}(a) and Figure~\ref{fig:parallel-usage}(b) shows the pseudocode of configuring Attention and FeedForward to conduct op-level parallelism, respectively.
\texttt{qkv\_mm}’s sharding strategy is \texttt{((2, 1), (1, 2))}, indicating that \texttt{x} is partitioned along the row (\textit{batch} or \textit{data}) dimension into 2 slices, while \texttt{q\_w}, \texttt{k\_w} and \texttt{v\_w} are partitioned along the column dimension. Since the device number is 4 here, each device holds a distinct pair of a \texttt{x}'s slice and a \texttt{q\_w}'s (\texttt{k\_w}'s and \texttt{v\_w}'s) slice. \texttt{matmul}’s sharding strategy is \texttt{((2, 2), (2, 1))}, where the contracting dimension is partitioned, thus an \texttt{AllReduce} is needed here to perform the operation. Likewise, another \texttt{AllReduce} is needed in Figure~\ref{fig:parallel-usage}(b)'s \texttt{matmul2}. Auto-parallel can find such needed operators.  
Furthermore, the \textit{tensor redistribution} is designed to automatically find the transformation (a list of operators) between any two inconsistent distributed tensor layouts with minimum communication cost, and then the operators are inserted into the data flow graph. The sharding strategy of \texttt{batch\_mm} in Figure~\ref{fig:parallel-usage}(a) corresponds to splitting the batch and head dimension.

Figure~\ref{fig:parallel-usage}(d) shows the pseudocode of conducting pipeline parallelism in MindSpore. The number of stages is configured as 2, and the number of devices is 8. Thus, 4 devices together perform each stage. The \texttt{layer1} is configured to be the stage 0, thus replicated on 4 devices. Likewise, \texttt{layer2} is replicated on the other 4 devices. If combined with Figure~\ref{fig:parallel-usage}(a) and Figure~\ref{fig:parallel-usage}(b), the desired parallelization strategy is obtained to \MODEL.\footnote{The stategy of optimizer parallelism is hidden in how batch dimension is split in the configuration. We omit the configuration for rematerialization here.} \texttt{Send} and \texttt{Receive} are inferred to communicate the activation output from stage 0 to stage 1, and then are automatically inserted into the data flow graphs on two stages, respectively.

In the future, we will: a) develop a cost model and a parallelization strategy searching algorithm for all parallelism dimensions in order to completely liberate developers from the underlying parallel-related works; b) support the heterogeneous-parallelism to offload a part of tensors and the corresponding computations to the host CPU to accelerate the training; c) use Sparse Attention to speedup the computation.

All training and inference jobs are run on the ModelArts\footnote{\url{https://www.huaweicloud.com/product/modelarts.html}} platform, which manages the end-to-end workflows and provides the functionality of cluster scheduling for a job to acquire a hierarchical cluster.

\section{Experiments}\label{sec:experiments}

\subsection{Training Details}
\begin{figure*}[]
	\centering
	\includegraphics[width=0.9\textwidth]{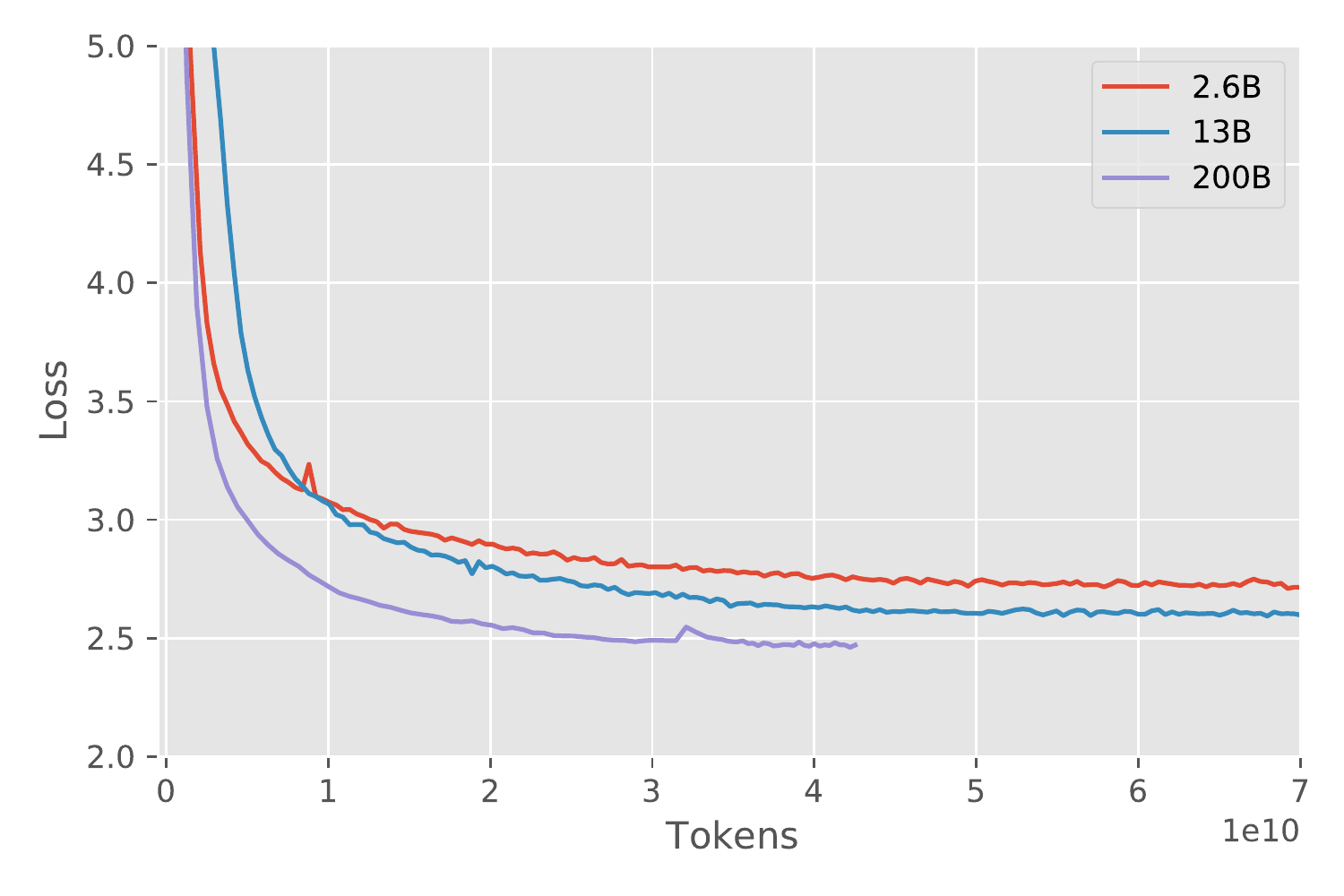}
	\caption{Training curves of three \MODEL\ models with different model sizes. The x-axis denotes the number of training tokens, which is measured as $training\_steps * batch\_size* sequence\_length$. The y-axis denotes the training loss.}
	\label{fig:loss_curve}
\end{figure*}
Our \MODEL\ models are developed under the Mindspore framework and are trained on a cluster of 2048 Ascend 910 AI processors. The detailed settings are shown in Table \ref{tab:training_details}. For the training of the 200B model, we use 2048 Ascend processors at the first phase and then switch to 1024 Ascends processors in the middle, in order to conduct other experiments using the rest of resources. The Byte Pair Encoding (BPE) is used as the tokenizer, and the vocabulary size are 40,000. The sequence length for the training data is set to 1024 for all the models. 

The curves of training loss for the \MODEL\ models are shown in Figure \ref{fig:loss_curve}. We adopt the number of training tokens as the x-axis since the batch size for the 200B model is not comparable to that of the 13B and 2.6B models. The loss of 200B model converges to around 2.49, while the losses of 13B and 2.6B models converge to 2.58 and 2.64 respectively. From the training curves, we can observed that the losses are still decreasing by the end of training, which indicates that our \MODEL\ model are still under-trained, and may have great potential to improve. We also evaluate the perplexity of our \MODEL\ models on the validation set, which is randomly sampled from the Common Crawl dataset. The results in Table \ref{tab:vad_ppl} show that \MODEL\ models with larger parameters sizes achieve smaller perplexity values, indicating that larger \MODEL\ models are better language models.
\begin{table*}[]
\centering
\caption{The detailed settings for training \MODEL\ models.}
\label{tab:training_details}
\begin{tabular}{l|l|l|l|l|l} \hline
Models                & \#Training Steps       & \#Ascend processors & Adam Betas & Learning Rate    &Weight Decay   \\ \hline
\MODEL\ 2.6B                  &  0$\sim$70,000                    &  512                 &    $\beta_1$=0.9 ,$\beta_2$=0.999        & 1e-4 & 0.01 \\ \hline
\MODEL\ 13B                   & 0$\sim$84,000        & 1024              & $\beta_1$=0.9 ,$\beta_2$=0.98       & 5e-5          & 0.01            \\ \hline
\multirow{2}{*}{\MODEL\ 200B} & 0$\sim$130,000       & 2048              & \multirow{2}{*}{$\beta_1$=0.9 ,$\beta_2$=0.95}        & \multirow{2}{*}{2e-5}  &           \multirow{2}{*}{0.1}       \\
                      & 130,000$\sim$260,000 & 1024              &        &   &                  \\ \hline
\end{tabular}
\end{table*}

\begin{table}[]
\centering
\caption{The validation perplexity of the \MODEL\ models.}
\label{tab:vad_ppl}
\begin{tabular}{l|c}
\hline
Models & Validation PPL \\ \hline
\MODEL\ 2.6B   & 19.33          \\ 
\MODEL\ 13B    & 17.69          \\
\MODEL\ 200B   & 15.59          \\\hline
\end{tabular}
\end{table}

\subsection{Task Description}
In this section, we evaluate our models on a broad spectrum of natural language processing tasks. Similar to the GPT-3 ~\cite{brown2020GPT3}, the experiments are conducted under three learning settings, i.e., zero-shot, one-shot, and few-shot, without any finetuning. For each task, we evaluate the models with the test sets when publicly available. Otherwise, we use the development sets instead. For some tasks with a very large test set or development set, we randomly sample a subset from the dataset in the experiments to reduce the computational cost. The evaluation datasets are classified into 7 categories by the task similarities, and we describe each category as follows.

\textbf{Cloze and completion tasks}, including WPLC, CHID ~\cite{CHID}, PD\&CFT ~\cite{PDCFT}, CMRC2017 ~\cite{CMRC2017}, and CMRC2019 ~\cite{CMRC2019}. 
Chinese WPLC (Word Prediciton with Long Context) is a dataset created to test the ability to model long-range dependencies, similar to the LAMBADA dataset ~\cite{LAMBADA} for English. The CHID (Chinese IDiom dataset) requires the model to identify the ground-truth idiom from 10 candidate idioms. The PD\&CFT task requires the model to predict the mask words in sentences derived from People's Daily (PD) news dataset and Children's Fairy Tale (CFT) dataset. The CMRC2017 (Chinese Machine Reading Comprehension) task contains two different sub-task: cloze-style task and user query reading comprehension task, among which we only evaluate our models on the cloze-style task. While the aforementioned tasks are word-level tasks, the CMRC2019 is a sentence cloze-style dataset that involves filling the right sentence from several candidate sentences into the passage. For the CMRC2019 and the CHID, a list of candidate choices are provided, making them classification tasks, while for WPLC, CMRC2017 and PD\&CFT, the models need to generate the answer as no candidate choices are given. Accuracy metric is employed for evaluating the cloze-style tasks.

\textbf{Reading comprehension tasks}, including CMRC2018 ~\cite{CMRC2018}, DRCD ~\cite{DRCD}, and DuReader ~\cite{DuReader}. These are all span-extraction tasks originally. That is, given a passage as context and a question, the models need to extract a text span from the passage which contains the correct answer to the question. The evaluation metrics, including F1 and exact match (EM), measure the similarity between the predicted span and the ground-truth text span. Instead of span-extraction, we formulate these task as generation tasks where the models generate the texts directly. The similarity between the generated text span and the ground-truth text span is evaluated. Note that for the DuReader task, we select the Zhidao subset for evaluation in our experiment.


\textbf{Closed-book question answering (QA) tasks}, including WebQA ~\cite{WebQA}. We follow the same closed-book setting in GPT-3 ~\cite{brown2020GPT3}, where the models are not allowed to access any external knowledge when answering open-domain factoid questions about broad factual knowledge. 

\textbf{Winograd-Style tasks}, including CLUEWSC2020 ~\cite{CLUE}. CLUEWSC2020 is a Chinese Winograd Schema Challenge dataset, which is an anaphora/coreference resolution task. In practice, we convert the task into a multiple-choice classification problem.

\textbf{Common sense reasoning tasks}, including C$^{3}$ ~\cite{CLUE}. C$^{3}$ is a free-form multiple-choice reading comprehension dataset which can benefit from common sense reasoning. Different from the extraction-based reading comprehension tasks, the answers to of C$^{3}$ questions cannot be directly found in the given context. Therefore, we use it to evaluate the common sense reasoning ability of the models.

\textbf{Natural language inference (NLI) tasks}, including Chinese Multi-Genre NLI (CMNLI) and Original Chinese Natural Language Inference (OCNLI) ~\cite{CLUE}. The NLI tasks require the model to identify the relation between two sentences, either entailment, neutral or contradiction. We formulate these tasks as three-class classification problems.

\textbf{Text classification tasks}, including TouTiao Text Classification for News Titles (TNEWS), IFLYTEK app description classification (IFLYTEK), Ant Financial Question Matching Corpus (AFQMC), and Chinese Scientific Literature (CSL) ~\cite{CLUE}.
These text classification tasks covers broad domains of text, including news, applications, financial text, scientific text. For the TNEWS and IFLYTEK tasks, there are 15 and 119 categories originally. However, we randomly sample three candidates as negative labels for each instance and perform 4-class classification. The reason is that the computational cost of our perplexity-based classification method increases linearly to the total number of candidate categories, which will be described in the next section.

\subsection{Evaluation Details}
The tasks can be generally classified into two-categories: classification tasks and generation tasks.  For the classification tasks, we resolve the task as perplexity comparison tasks.
For some tasks, the samples needs to be filled into a tailor-designed template as the input to the models. The templates for each task are described in Table \ref{tab:prompt_setting}, where "/" means the task does not involve a template. The decoding strategies for these text generation tasks are described in Table \ref{tab:decoding_strategies}.

\begin{CJK}{UTF8}{gbsn}
\begin{table}
\centering
\renewcommand\arraystretch{1.25}
\caption{The input\&prompt template for each task.}
\label{tab:prompt_setting}
\resizebox{\textwidth}{!}{
\begin{tabular}{llll} 
\hline

\textbf{Task}          & \textbf{Dataset}        & \textbf{Input\&Prompt} \\

\hline
Cloze and completion   & \begin{tabular}[c]{@{}l@{}} WPLC\\ CHID\\ PD\&CFT\\ CMRC2017\\ CMRC2019 \end{tabular}        
                       & \begin{tabular}[c]{@{}l@{}} /\\ /\\ /\\/\\ /\\ / \end{tabular}\\
                       
\specialrule{0em}{2pt}{2pt} 

Reading comprehension  & \begin{tabular}[c]{@{}l@{}} CMRC2018\\ DRCD\\ DuReader \end{tabular}
                       & \begin{tabular}[c]{@{}l@{}} 阅读文章：\$Document\textbackslash{}n问：\$Question\textbackslash{}n答：(Read document: \$Document\textbackslash{}nQuestion：\$Question\textbackslash{}nAnswer: )\\ 阅读文章：\$Document\textbackslash{}n问：\$Question\textbackslash{}n答：(Read document: \$Document\textbackslash{}nQuestion：\$Question\textbackslash{}nAnswer: )\\ 阅读文章：\$Document\textbackslash{}n问：\$Question\textbackslash{}n答：(Read document: \$Document\textbackslash{}nQuestion：\$Question\textbackslash{}nAnswer: \end{tabular}\\
                       
\specialrule{0em}{2pt}{2pt}

Closed book QA         & WebQA 
                       & 问：\$Question\textbackslash{}n答：(Question：\$Question\textbackslash{}nAnswer: )\\
                       
\specialrule{0em}{2pt}{2pt}    

Winograd-Style         & CLUEWSC2020
                       & /\\
                       
\specialrule{0em}{2pt}{2pt} 

Common sense reasoning & C\textsuperscript{3} 
                       & 问: \$Question\textbackslash{}n答:\$Choice\textbackslash{}n该答案来自对话: \$Passage (Question: \$Question\textbackslash{}nAnswer:\$Choice\textbackslash{}nAnswer from dialogue: \$Passage)\\

\specialrule{0em}{2pt}{2pt} 
                       
NLI                    & \begin{tabular}[c]{@{}l@{}} CMNLI\\ OCNLI \end{tabular}
                       & \begin{tabular}[c]{@{}l@{}} \$S1?对/或许/错，\$S2  (\$S1?Yes/Maybe/No, \$S2)\\ \$S1?对/或许/错，\$S2  (\$S1?Yes/Maybe/No, \$S2) \end{tabular}\\

\specialrule{0em}{2pt}{2pt} 

Text classification    & \begin{tabular}[c]{@{}l@{}} TNEWS\\ IFLYTEK\\ AFQMC\\ CSL \end{tabular}
                       & \begin{tabular}[c]{@{}l@{}} 这是关于\$label的文章：\$passage (This passage is about\$label: \$passage)\\ 这是关于\$label的应用程序：\$passage (This application is about: \$passage)\\ 下面两个句子语义相同/不同：\$S1。\$S2 (The following two sentences have the same/different semantics: \$S1. \$S2)\\ 摘要：\$passage，关键词：\$keyword是/不是真实关键词 
                      (Abstract: \$passage, keywords: \$keyword True/False keywords) \end{tabular}\\
\hline
\end{tabular}}
\end{table}
\end{CJK}

\begin{table}
\centering
\renewcommand\arraystretch{1.25}
\caption{The decoding strategies for text generation tasks.}
\label{tab:decoding_strategies}
\begin{tabular}{llll} 
\hline

Task                   & Dataset        & Decoding strategies \\

\hline
Cloze and completion   & \begin{tabular}[c]{@{}l@{}} WPLC \\ PD\&CFT \\ CMRC2017 \end{tabular} 
                       & \begin{tabular}[c]{@{}l@{}} top-k, k=1  \\ top-k, k=1,temperature=0.9  \\ top-p, p=0.9, temperature=1 \end{tabular}\\

Reading comprehension  & \begin{tabular}[c]{@{}l@{}} CMRC2018 \\ DRCD \\ DuReader \end{tabular}
                       & \begin{tabular}[c]{@{}l@{}} top-p, p=0.8, temperature=0.8 \\ top-p, p=0.8, temperature=0.8 \\ top-p, p=0.9, temperature=0.7 \end{tabular} \\
                       
Closed book QA         & WebQA 
                       & top-k, k=5 \\
                       
\hline
\end{tabular}
\end{table}

\subsubsection{Generation method} The generation tasks include word-level generation tasks and sentence-level generation tasks. Since our \MODEL\ models are autoregressive language models capable of text generation, the generation tasks can be solved naturally by simply generating the answers. For the cloze tasks such as WPLC, PD\&CFT, and CMRC2017, the prompts are the context before the positions to be predicted. For the reading comprehension tasks and closed book QA tasks, templates are designed if necessary. For example, in the reading comprehension tasks, the sample is filled into a template \textit{Reading document : \$Document Question: \$Question Answer:}, which serves as the prompt for the model to generate the answer.

\begin{figure}[h!]
\centering
\includegraphics[scale=0.6]{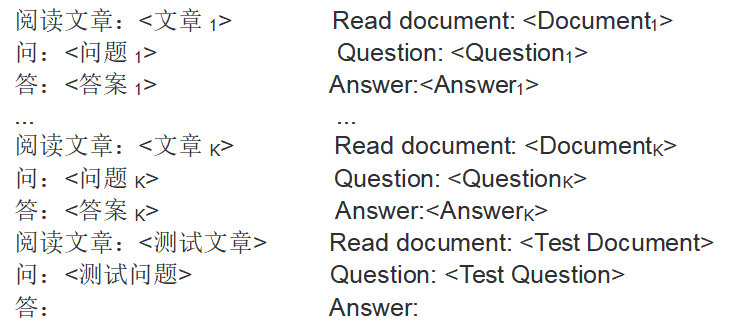}
\caption{A prompt for generation task of CMRC2018}
\label{fig:rs_prompt_demo}
\end{figure}

As in GPT-3, the few-shot task is designed as in-context learning, where $K$ prompts are concatenated one by one. The first $K-1$ prompts contain the ground truth answer while the last prompt is the sample we want to predict. An example for CMRC2018 task is shown in Figure \ref{fig:rs_prompt_demo}

\subsubsection{Perplexity-based method} The perplexity-based method solves the classifications tasks. For each pair of <text, label>, an input will be generated automatically according to a pre-designed criteria, as shown in Table \ref{tab:prompt_setting}. The sequence generated by the template will be fed into the model and a perplexity value will be computed. The label associated with the smallest perplexity value will be considered as the predicted label for this passage.

\begin{figure}[h!]
\centering
\includegraphics[scale=0.6]{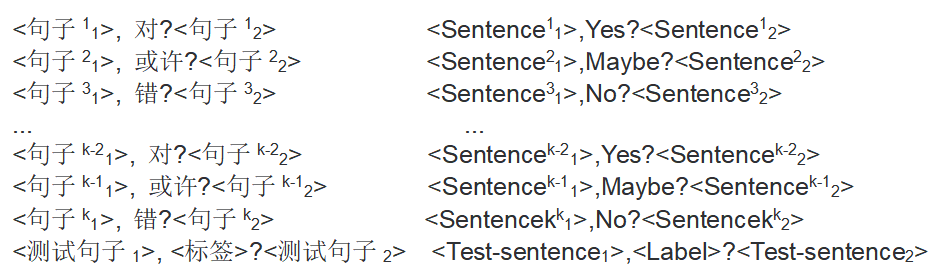}
\caption{Prompt for perplexity-based tasks of OCNLI}
\label{fig:ppl_prompt_demo}
\end{figure}

We also employ the in-context learning strategy for solving few-shot tasks. An example for few-shot OCNLI task is shown in Figure \ref{fig:ppl_prompt_demo}.

\subsection{Results}
Table \ref{tab:VS26B} compares \MODEL\ 2.6B with CPM~\cite{CPM2020}~\footnote{\url{https://github.com/TsinghuaAI/CPM-Generate}}, a recently released generative Chinese PLM with 2.6B parameters, on 16 downstream tasks in Chinese. \MODEL\ 2.6B achieves higher performance compared to CPM 2.6B on more than 11 tasks in zero-shot setting, 12 tasks on the one-shot setting, and 14 tasks on the few-shot setting. In general, the experimental results indicate that \MODEL\ 2.6B achieves higher in-context learning ability over CPM 2.6B, especially for few-shot learning and generation-tasks. Regarding generation-tasks, \MODEL\ 2.6B outperforms CPM 2.6B with an improvement of 6 points on average. To be more specific, \MODEL\ 2.6B surpasses CPM 2.6B with 5 points in scores for both reading comprehension and closed-book QA tasks, 7 points in scores for cloze (without choices) tasks respectively. Regarding perplexity-tasks, \MODEL is comparable to CPM 2.6B on natural language inference with CMNLI and OCNLI datasets, while it is slightly worse than CPM on classification tasks with TNEWS and IFLYTEK datasets. We suppose that the main factor that contributes to the different performance of CPM 2.6B and \MODEL\ 2.6B is the training data. We collect massive and diverse data from a wide range of sources, which allows our \MODEL\ model to handle more diverse tasks. 

\begin{table}
\centering
\renewcommand\arraystretch{1.25}
\caption{Performance comparison of CPM 2.6B v.s. \MODEL\ 2.6B on few-shot NLP tasks.}
\label{tab:VS26B}
\setlength{\tabcolsep}{1.5mm}
\begin{threeparttable}
\resizebox{\textwidth}{33mm}{\begin{tabular}{c|c|c|c|c c|c c|c c c} 
\hline


\multicolumn{1}{c|}{}     & \multicolumn{1}{c|}{}       & \multicolumn{1}{c|}{}        & \multicolumn{1}{c|}{}           & \multicolumn{2}{c|}{Zero-Shot}                                 & \multicolumn{2}{c|}{One-Shot}                                  & \multicolumn{3}{c}{Few-Shot}                                                          \\
\multicolumn{1}{c|}{Dataset} & \multicolumn{1}{c|}{Method} & \multicolumn{1}{c|}{Metrics} & \multicolumn{1}{c|}{Task Types} & \multicolumn{1}{c}{CPM 2.6B} & \multicolumn{1}{c|}{\MODEL\ 2.6B} & \multicolumn{1}{c}{CPM 2.6B} & \multicolumn{1}{c|}{\MODEL\ 2.6B} & \multicolumn{1}{c}{\#Shot($K$)} & \multicolumn{1}{c}{CPM 2.6B} & \multicolumn{1}{c}{\MODEL\ 2.6B} \\

\hline
CMRC2018 & Generation &  Em/F1  & Read Comprehension & 0.59/10.12 & \textbf{1.21/16.647} & 1.71/11.29 & \textbf{2.49/18.57} & Dynamic & 3.11/14.64 & \textbf{5.68/23.22} \\

\hline
DRCD & Generation & Em/F1 & Read Comprehension & 0/4.62 & \textbf{0.8/9.99} & 0.22/5.17 & \textbf{2.47/12.48}  & Dynamic & 0.15/7.14 & \textbf{5.31/18.29} \\

\hline
DuReader & Generation &  Rouge-1  & Read Comprehension & 16.63 & \textbf{21.07} & 16.42 &	\textbf{20.18} & 6,6 & 17.85 &	\textbf{21.43} \\

\hline
WebQA & Generation & 	Em/f1  & Closed-Book QA	& 6/12.59	& \textbf{6/16.32}	& 6/11.82	& \textbf{12/23.39} & 8,8 &	4/12.23 &	\textbf{24/33.94} \\

\hline
PD-CFT & Generation & 	Acc  &	Cloze(without choices)	& 35.73/38.99 &	\textbf{38.47/42.39} &	33.3/39.73 &	\textbf{38.8/41.61} & 3,3 &	32.03/39.84 & \textbf{39.07/42.05} \\

\hline
CMRC2017	& Generation & 	Acc & Cloze(without choices)	& 24.60 	& \textbf{37.83} &	25.40 &	\textbf{38.00} 	& 3,3 & 23.50 &	\textbf{36.33}  \\

\hline
CHID	& PPL & 	Acc	 & Cloze(multi-choices)	& 68.62 	& \textbf{68.73} &	67.91 &	\textbf{68.16} & 3,3 &	\textbf{66.82} &	66.56  \\
\hline
CMRC2019 & PPL &  Acc & Cloze (multi-choices) & 47.69 & \textbf{61.93} & 47.99 & \textbf{61.54}  & 2,2 &  47.20 & \textbf{62.42} \\
\hline
CMNLI &	PPL &	Acc  &	Natural Language Inference &	49.10 & 	\textbf{50.20} & 	47.56 & 	\textbf{49.54} & 6,12 & 	49.29 & 	\textbf{51.17} \\
\hline
OCNLI &	PPL &	Acc  &	Natural Language Inference &	\textbf{44.20} & 	42.61 & 	\textbf{44.30} & 	44.00 &  3,6	& 44.00 & 	\textbf{46.78} \\
\hline
TNEWS &	PPL &		Acc &	Text classification &	\textbf{65.44} & 	60.95 & 	\textbf{69.50} & 	57.95 &  6,6 & 	\textbf{70.17} & 	63.62 \\
\hline
IFLYTEK &	PPL &	Acc  & Text classification &	68.91 & 	\textbf{74.26} & 	\textbf{79.84} & 	79.03 & 3,3 & 	\textbf{83.99} & 	80.15 \\
\hline
AFQMC &	PPL &	Acc  & Sentence Pair Similarity &	\textbf{66.34} & 	59.29 & 	39.70 & 	\textbf{64.62} & 4,4 & 	38.29 & 	\textbf{69.00} \\
\hline
CSL &	PPL &Acc  &	Keyword  Recognition &	\textbf{52.30} & 	50.50 & 	\textbf{51.20} & 	50.90 & 10,10 & 	50.50 & 	\textbf{52.00} \\
\hline
CLUEWSC2020 &	PPL &	 Acc  &	WSC &	\textbf{73.684} & 	73.36 & 	73.684 & 	\textbf{75.33} &	14,14  & 	70.065 & 	\textbf{72.70} \\
\hline
C\textsuperscript{3} &	PPL &		Acc  &	Common Sense Reasoning &	49.81 & 	\textbf{53.42} & 	51.43 & 	\textbf{52.82}  &	3,3 & 	51.60 & 	\textbf{53.64} \\
\hline

\end{tabular}}
    \end{threeparttable}
\end{table}

\begin{table}
\centering
\renewcommand\arraystretch{1.25}
\caption{Performance comparison of \MODEL\ 2.6B v.s. \MODEL\ 13B on few-shot NLP tasks.}
\label{tab:VS13B}
\setlength{\tabcolsep}{1.5mm}
\begin{threeparttable}
\resizebox{\textwidth}{33mm}{\begin{tabular}{c|c|c|c|c c|c c|c c c} 
\hline

\multicolumn{1}{c|}{}     & \multicolumn{1}{c|}{}       & \multicolumn{1}{c|}{}        & \multicolumn{1}{c|}{}           & \multicolumn{2}{c|}{Zero-Shot}                                 & \multicolumn{2}{c|}{One-Shot}                                  & \multicolumn{3}{c}{Few-Shot}                                                          \\
\multicolumn{1}{c|}{Dataset} & \multicolumn{1}{c|}{Method} & \multicolumn{1}{c|}{Metrics} & \multicolumn{1}{c|}{Task Types} & \multicolumn{1}{c}{\MODEL\ 2.6B} & \multicolumn{1}{c|}{\MODEL\ 13B} & \multicolumn{1}{c}{\MODEL\ 2.6B} & \multicolumn{1}{c|}{\MODEL\ 13B} & \multicolumn{1}{c}{\#Shot($K$)} & \multicolumn{1}{c}{\MODEL\ 2.6B} & \multicolumn{1}{c}{\MODEL\ 13B} \\
\hline
CMRC2018 & Generation & Em/F1  & Read Comprehension & 1.21/16.65 & \textbf{1.46/19.28} & 2.49/18.57 & \textbf{3.76/21.46} & Dynamic & 5.68/23.22 & \textbf{9.76/29.23} \\
\hline
DRCD & Generation & Em/F1   & Read Comprehension & \textbf{0.8}/9.99	 & 0.66/\textbf{10.55} & 2.47/12.48 & \textbf{4.22/15.01} & Dynamic & 5.31/18.29 & \textbf{9.09/23.46}\\
\hline
DuReader & Generation & Rouge-1  & Read Comprehension & 21.07  & \textbf{24.46}  & 20.18  & \textbf{25.99}  & 6,6 & 21.43  & \textbf{27.67}  \\
\hline
WebQA & Generation & Em/f1  & Closed-Book QA	& 4.43/13.71 & \textbf{5.13/14.47} & 10.22/20.56 & \textbf{13.43/24.52} & 8,8 & 23.71/33.81 & \textbf{31.18/41.21} \\
\hline
PD-CFT & Generation & 	Acc &	Cloze(without choices)	& 38.47/42.39  & \textbf{43.86/46.60}  & 38.8/41.61  & \textbf{40.97/45.42} & 3,3  & 39.07/42.05  & \textbf{41.13/45.86}  \\
\hline
CMRC2017	& Generation & 	Acc & Cloze(without choices)	& 37.83  & \textbf{38.90}  & 38.00  & \textbf{38.40}  	& 3,3 & 36.33  & \textbf{37.86}   \\
\hline
 CHID	& PPL &	Acc	 & Cloze(multi-choices)	& 68.73  & \textbf{70.64}  & 68.16  & \textbf{70.05} & 3,3 & 66.56  & \textbf{70.91}   \\
\hline
CMRC2019 & PPL &  Acc	& Cloze (multi-choices) & 68.22  & \textbf{70.54}  & 68.05  & \textbf{70.02}  & 2,2 & 66.26  & \textbf{71.28}  \\
\hline
CMNLI & PPL & Acc &	Natural Language Inference &	\textbf{50.20}  & 48.44  & \textbf{49.54}  & 46.81 & 6,12  & \textbf{51.17}  & 46.18  \\
\hline
OCNLI & PPL & Acc &	Natural Language Inference &	\textbf{42.61}  & 41.53  & 44.00  & \textbf{44.10} & 3,6  & \textbf{46.78}  & 46.44  \\
\hline
TNEWS & PPL &	Acc &	Text classification &	 \textbf{60.95}  & 60.26  & 57.95  & \textbf{63.83}  & 6,6  & 63.62  & \textbf{65.17}  \\
\hline
IFLYTEK & PPL &	Acc &	Text classification &	\textbf{74.26}  & 73.80  & \textbf{79.03}  & 78.95 & 3,3  & 80.15  & \textbf{80.34}  \\
\hline
AFQMC & PPL &	Acc &	Sentence Pair Similarity &	59.29  & \textbf{65.76}  & \textbf{64.62}  & 63.55 & 4,4  & \textbf{69.00}  & 68.91  \\
\hline
CSL & PPL &	Acc&	Keyword  Recognition &	\textbf{50.50}  & 49.30  & \textbf{50.90}  & 50.20  & 10,10  & 52.00  & \textbf{55.70}  \\
\hline
CLUEWSC2020 & PPL &	Acc  &	WSC &	73.36  & \textbf{75.00}  & \textbf{75.33}  & 75.00 & 14,14  & 72.70  & \textbf{78.62} \\
\hline
C\textsuperscript{3} & PPL &	Acc&	Common Sense Reasoning &	53.42  & \textbf{54.47}  & 52.82  & \textbf{53.92}  & 3,3  & 53.64  & \textbf{54.58} \\
\hline
WPLC &	PPL &	ppl &	Chinese WPLC &	16.70  & \textbf{19.18}  & -  & - & - & - & - \\
\hline

\end{tabular}}
    \end{threeparttable}

\end{table}

Table \ref{tab:VS13B} compares \MODEL\ 13B with \MODEL\ 2.6B. \MODEL\ 13B outperforms \MODEL\ 2.6B on all generation-tasks and most of the perplexity-tasks. Regarding CMRC2018, DRCD and WebQA tasks of \MODEL\ 13B, the few-shot performance surpasses zero-shot by more than 10 points, demonstrating that \MODEL\ 13B has superior in-context learning ability. 
 \MODEL\ 13B outperforms \MODEL\ 2.6B  with an improvement of 3 points on average. To be more specific, \MODEL\ 13B surpasses \MODEL2.6B  with 4 points for both reading comprehension and closed-book QA tasks, 2 points for cloze (without choices) tasks respectively. Regarding the NLI tasks, the 13B model performs worse than the 2.6B model, which is consistent with the observations in GPT-3. 
 Overall, the comparison results between \MODEL\ 13B with \MODEL\ 2.6B demostrate that a larger scale of pretrained model generally improves the performance on few-shot learning tasks. 

\subsection{Natural Language Generation Examples}
We evaluate the generation capabilities of \MODEL\ 200B on various text generation scenarios. We show some of the examples in this section. We do not conduct any post-editing to the generated text, except that we truncate the generated text when the model does not stop generation at a reasonable point. Among the scenarios we have tested, we find that our \MODEL\ model is particularly good at poetry\&duilian generation, text summarization, dialog generation, and fiction generation, where roughly 90\% of the generated examples are acceptable to human. We believe there are certainly more applications for \MODEL\ models to explore in the future.
\begin{table}[]
\centering
\caption{An example of the reading comprehension by \MODEL\ model.}
\begin{tabular}{l|l} \hline
\multicolumn{2}{c}{Reading Comprehension } \\ \hline
Prompt                   &    
            
            \parbox{0.8\textwidth}{%
        \begin{CJK}{UTF8}{gbsn}
        阅读文章：株洲北站全称广州铁路（集团）公司株洲北火车站。除站场主体，另外管辖湘潭站、湘潭东站和三个卫星站，田心站、白马垅站、十里冲站，以及原株洲车站货房。车站办理编组、客运、货运业务。车站机关地址：湖南省株洲市石峰区北站路236号，邮编412001。株洲北站位于湖南省株洲市区东北部，地处中南路网，是京广铁路、沪昆铁路两大铁路干线的交汇处，属双向纵列式三级七场路网性编组站。车站等级为特等站，按技术作业性质为编组站，按业务性质为客货运站，是株洲铁路枢纽的主要组成部分，主要办理京广、沪昆两大干线四个方向货物列车的到发、解编作业以及各方向旅客列车的通过作业。每天办理大量的中转车流作业，并有大量的本地车流产生和集散，在路网车流的组织中占有十分重要的地位，是沟通华东、华南、西南和北方的交通要道，任务艰巨，作业繁忙。此外，株洲北站还有连接石峰区喻家坪工业站的专用线。株洲北站的前身是田心车站。
请问：株洲北站的前身是哪个车站？回答：
        \end{CJK}

        }            \\ \hline
Generation               &  \begin{CJK}{UTF8}{gbsn}田心车站\end{CJK} \\ \hline
\end{tabular}
\end{table}

\begin{table}[]
\centering
\caption{An example of open domain QA by \MODEL\ model. The second example implies that \MODEL\ model does not simply act like a retrieval model to answer open domain questions, but can understand and decompose the questions and generated answers accordingly.}
\begin{tabular}{l|l} \hline
\multicolumn{2}{c}{Open Domain Question Answering} \\ \hline
Prompt                   &    
            
            \parbox{0.8\textwidth}{%
        \begin{CJK}{UTF8}{gbsn}
       中国的首都是哪里？
        \end{CJK}
        }            \\ 
Generation               &  \begin{CJK}{UTF8}{gbsn}中国的首都是北京。\end{CJK} \\ \hline
Prompt                   &  
            
            \parbox{0.8\textwidth}{%
        \begin{CJK}{UTF8}{gbsn}
       中国、美国、日本和澳大利亚的首都分别是哪里？
        \end{CJK}
        }            \\ 
Generation               &  \begin{CJK}{UTF8}{gbsn}中国的首都是北京，美国的首都是华盛顿，日本的首都是东京，澳大利亚的首都是堪培拉。\end{CJK} \\ \hline

Prompt                   &   
            
            \parbox{0.8\textwidth}{%
        \begin{CJK}{UTF8}{gbsn}
       五一川西行程规划

        \end{CJK}
        }            \\ 
Generation               & 
            
            \parbox{0.8\textwidth}{ \begin{CJK}{UTF8}{gbsn}五一川西旅游活动主要集中在绵阳、成都、雅安、甘孜4市（州），主要景区有大九寨、新都桥、丹巴、海螺沟、泸定、康定、成都新会展中心。\end{CJK}} \\ \hline
\end{tabular}
\end{table}

\begin{table}[]
\centering
\caption{\MODEL\ model can generate Gushi and Duilian without finetuning on the domain-specific data.}
\begin{tabular}{l|l} \hline
\multicolumn{2}{c}{Gushi \& Duilian Generation } \\ \hline
Prompt                   &   
            
            \parbox{0.8\textwidth}{%
        \begin{CJK}{UTF8}{gbsn}
以枯藤、老树、夕阳作一首诗：
        \end{CJK}

        }            \\ 
Generation               &  \begin{CJK}{UTF8}{gbsn}藤影参差绕蓝桥，野花压枝低袅风，山风习习观残雪，村舍依依麦已深。\end{CJK} \\ \hline

Prompt                   &   
            
            \parbox{0.8\textwidth}{%
        \begin{CJK}{UTF8}{gbsn}
上联：欢天喜地度佳节
        \end{CJK}

        }            \\ 
Generation               &  \begin{CJK}{UTF8}{gbsn}下联：喜气洋洋迎新年\end{CJK} \\ \hline

\end{tabular}
\end{table}

\begin{table}[]
\centering
\caption{Our \MODEL\ model is especially good at text summarization. We tried several latest news articles which are not in our training data. For more than 90\% of the input article, our \MODEL\ model can generate satisfactory results. Source of the prompt: \begin{CJK}{UTF8}{gbsn}https://new.qq.com/omn/20210421/20210421A079J800.html\end{CJK}}
\begin{tabular}{l|l} \hline
\multicolumn{2}{c}{Text Summarization } \\ \hline
Prompt                   &   
            
            \parbox{0.8\textwidth}{%
        \begin{CJK}{UTF8}{gbsn}
“据媒体报道，最初华为只决定做车联网。2012年，华为开始进行汽车相关研究，当时在2012实验室下成立车联网实验室，研究电动汽车技术。

2013年，华为分别展示过为汽车内部提供3GWifi热点的DA6810、应用于保险行业及车队管理的汽车在线诊断系统DA3100，以及车载通信模块产品“ME909T”。

当时，华为对外的说法是，“华为终端只会聚焦于自己擅长的车联网通信解决方案。华为愿景是做到车联网模块第一，业绩上做到50亿~60亿元的规模”。

2019年5月27日，任正非签发华为组织变动文件，批准成立智能汽车解决方案BU部门，隶属于ICT管理委员会管理。智能汽车解决方案BU之下分：智能座舱部门、自动驾驶部门、智能车云服务部门。到这个阶段，华为已经确立了自动驾驶部门。

2020年7月12日晚，华为消费者业务CEO余承东官宣了HUAWEI　HiCar车机系统，HiCar系统基于鸿蒙OS。HiCar可以利用移动设备的5G网络能力加强车机系统的数据交互能力，能够接入华为在AI、语音、计算机视觉等方面的能力。同时，能够调用车速、方向盘转角、档位模式、汽车环境光传感器在内的车身数据以及空调、车窗、喇叭等车身控制部件。

直到徐直军在2021年的分析师大会透露出一个重点信号，华为将汽车服务的重心定在自动驾驶软件上，才明确了华为的自动驾驶业务。据媒体报道，华为跟百度、小鹏、文远知行、驭势科技等一样，都是选择在乘用车领域重点布局。”的新闻摘要是
        \end{CJK}

        }            \\ \hline
Generation               &  \parbox{0.8\textwidth}{\begin{CJK}{UTF8}{gbsn}“华为在自动驾驶领域的布局，是基于华为在人工智能、大数据、云计算等方面的技术积累，以及华为在汽车领域的多年经验。”\end{CJK}} \\ \hline\hline

Prompt                   &   
            
            \parbox{0.8\textwidth}{%
        \begin{CJK}{UTF8}{gbsn}
“据媒体报道，最初华为只决定做车联网。2012年，华为开始进行汽车相关研究，当时在2012实验室下成立车联网实验室，研究电动汽车技术。

2013年，华为分别展示过为汽车内部提供3GWifi热点的DA6810、应用于保险行业及车队管理的汽车在线诊断系统DA3100，以及车载通信模块产品“ME909T”。

当时，华为对外的说法是，“华为终端只会聚焦于自己擅长的车联网通信解决方案。华为愿景是做到车联网模块第一，业绩上做到50亿~60亿元的规模”。

2019年5月27日，任正非签发华为组织变动文件，批准成立智能汽车解决方案BU部门，隶属于ICT管理委员会管理。智能汽车解决方案BU之下分：智能座舱部门、自动驾驶部门、智能车云服务部门。到这个阶段，华为已经确立了自动驾驶部门。

2020年7月12日晚，华为消费者业务CEO余承东官宣了HUAWEI　HiCar车机系统，HiCar系统基于鸿蒙OS。HiCar可以利用移动设备的5G网络能力加强车机系统的数据交互能力，能够接入华为在AI、语音、计算机视觉等方面的能力。同时，能够调用车速、方向盘转角、档位模式、汽车环境光传感器在内的车身数据以及空调、车窗、喇叭等车身控制部件。

直到徐直军在2021年的分析师大会透露出一个重点信号，华为将汽车服务的重心定在自动驾驶软件上，才明确了华为的自动驾驶业务。据媒体报道，华为跟百度、小鹏、文远知行、驭势科技等一样，都是选择在乘用车领域重点布局。”的新闻标题是《
        \end{CJK}

        }            \\ \hline
Generation               &  \parbox{0.8\textwidth}{\begin{CJK}{UTF8}{gbsn}华为在车联网业务方面有深厚积累，并将全面发力自动驾驶软件业务》\end{CJK}} \\ \hline
\end{tabular}
\end{table}

\begin{table}[]
\centering
\caption{Examples of QA with long answers by \MODEL. These questions are from Zhihu.com. Overall, the answers seems plausible, but still contain irrelevant or repetitive or illogical sentences.}
\begin{tabular}{l|l} \hline
\multicolumn{2}{c}{Article Generation} \\ \hline
Prompt                   &   
            
            \parbox{0.8\textwidth}{%
        \begin{CJK}{UTF8}{gbsn}
有哪些办法可以教育猫？
        \end{CJK}

        }            \\ \hline
Generation               &  \parbox{0.8\textwidth}{\begin{CJK}{UTF8}{gbsn}我觉得猫并不是不懂事的孩子，而是不懂事的孩子才会乱咬东西，所以，最好的办法就是给猫咪找个好的主人。楼主给它找家宠物店的话，在宠物店里买猫砂，不过，家里应该再放一个猫盆或者专用的宠物用盆。还有要提醒楼主一点的是，猫咪的牙齿问题也很容易使它们咬东西，所以，一定要给猫咪好好刷牙，定期的给猫咪洗澡，因为猫咪喜欢闻自己的身体的味道，所以，洗澡也是一种很好的办法。在给猫咪买吃的东西的时候，一定要注意，那些味道重的，可能就是猫咪误食的了，那就赶紧停止吃东西，或者把饭菜都吐出来，不然的话，就很可能误食一些它没有办法去辨识的东西，所以，为了防止这些东西误食，最好还是给猫咪买猫咪专用的，防止猫咪误食的。在给猫咪换洗干净的衣服或者是被子的时候，一定要用干净的猫砂盆，不要用热水，也不要用很烫的水，因为猫咪的身体素质很好，耐热能力也很强，所以，很容易伤到它们的身体。\end{CJK}} \\ \hline\hline

Prompt                   &   
            
            \parbox{0.8\textwidth}{%
        \begin{CJK}{UTF8}{gbsn}
猫可以聪明到什么程度？
        \end{CJK}

        }            \\ \hline
Generation               &  \parbox{0.8\textwidth}{\begin{CJK}{UTF8}{gbsn}流浪猫可以聪明到什么程度？我想够聪明的猫应该算是了。那么我们来看看，到底有多聪明。美国科学家研究发现，猫学会了同人类社会和动物社会的语言和相处的能力，这也是它能从孤僻的猫种中脱颖而出的原因。其中包括做过面部美容手术以及爱好钓鱼的猫。曾有这样一个实验，在野外的一个植物园，喂养了一批神秘猫。在吃了这些猫饲料后，发现这些猫学习能力、注意力以及对野外生存能力的提高幅度都要比没喂这些猫饲料的猫高出许多。另外，动物学家对猫行为的分析研究发现，猫善于察言观色，通过观察在野外工作的猫，能逐渐熟悉它的身体构造。观察者指出，猫能通过眼睛和嘴来收集情报，在它们与我们人共同生活的初期，就能把眼神、表情、姿势和其他习性反映给我们，而且它们能学会在距离不远的地方伏击猎物。而生活在现代社会的我们，早已学会了隐藏和保密。久而久之，我们培养起了对“猫“的感情依赖，“猫”也自然而然成了我们生活中不可缺少的一部分。正所谓：猫能够聪明到什么程度？\end{CJK}} \\ \hline
\end{tabular}
\end{table}

\begin{table}[]
\centering
\caption{Our \MODEL\ model is good at dialog generation. Given a shot conversation as prompt, the model generates several rounds of conversations automatically. This implies that \MODEL\ model has potentials to assist humans in automatic reply in customer service, or design dialogues for NPC (Non-Player Character) in games.\begin{CJK}{UTF8}{gbsn}
Source of Prompt: https://www.zhihu.com/question/313316339/answer/623572659. Author：风吹走的名字.
\end{CJK}}
\begin{tabular}{l|l} \hline
\multicolumn{2}{c}{Game Dialog Generation} \\ \hline
Prompt                   &  
            
            \parbox{0.8\textwidth}{%
        \begin{CJK}{UTF8}{gbsn}
书生：羌笛何须怨杨柳，春风不度玉门关。\\
飞云：（这诗怎么这么耳熟？且过去跟他聊聊如何。）\\
书生：小兄弟，要不要一起喝一杯？\\
飞云：你请我呀？你若是请我，我便和你喝一杯；你若不请我，我便一个人去喝。\\
书生：小兄弟，看你年纪轻轻，不至于这么势利吧？\\
飞云：势利如何说起？你请我喝了酒，我自然要记你的人情；我欠了你的人情，日后你有求于我，我自然要报答你，这样一来大家都不吃亏，不过你的好处要在后头得到而已。\\
书生：小兄弟言之有理，不过在下有一点不太明白，那就是为什么你不愿意先请在下喝了酒待日后在下回报于你呢？
        \end{CJK}

        }            \\ \hline
Generation1               &  \parbox{0.8\textwidth}{\begin{CJK}{UTF8}{gbsn}你若是愿意先请我喝了酒，待日后我回报于你，你便有了收获，你岂不是更先欠下我的人情了吗？\\
飞云：这是自然的，我也说过了，我的确是不喜欢欠别人的人情，更不喜欢别人欠我的人情，那便是把好处在后头得到了。\\
书生：小兄弟，我是一介书生，你在座的各位都是我的衣食父母，你们如果都像我一样想要什么，要不到了我，那你们会像我一样被饿死、被寒冷冻死吗？\\
飞云：那是因为在座的各位都是饥寒交迫之人，都是饿得骨瘦如柴之人，而我却是经常吃饱喝足、锦衣玉食、养尊处优之人，这正是所谓的乐施善心，积德行善之人。\\
书生：小兄弟，你是不是因为这样才不愿意欠下我的人情？\\
飞云：不是，因为我知道，你如果不愿意先请我喝了酒待日后我回报于你，那么我便不能报答你;而我不愿意欠别人的人情，那我便会将自己的美好的善行全部都奉献给你，以报答你对我的关怀和帮助。\\
书生：小兄弟，不知你用什么方法找到这么多可敬可爱的人们的？\\
飞云：呵呵，说来惭愧，我找他们可不是什么难事，我认识他们的时候，他们也还没有这么大的势力，我只是为了帮他们找到一条可以生存的路，从他们那里换到了一些粮食。只是那时候我对他们的帮助确实太少了，所以最后他们都没有答应我的要求，也没有向我索要东西，甚至于没有等到我的帮助便饿死了。\\
书生：小兄弟，我看你年纪轻轻，怎么会有如此崇高的思想境界呢？又是从哪里学来的呢？你说来听听。\\
飞云：这倒说不上，只不过是我发现了很多常人做不到的事情，所以就学会了很多常人不会去做的事情，所谓饮水思源，这便是其中的一个原因。\\
书生：你说的是不是为了一些自己不知道的事物，而发奋图强？\\
飞云：是啊，我一直都\end{CJK}} \\ \hline

Generation2               &  \parbox{0.8\textwidth}{\begin{CJK}{UTF8}{gbsn}飞云：咳，年轻人说的话要有选择性，我既然说了我先请你，你却又说一个人喝一杯，可见我是有目的的，要是小兄弟真是不肯给在下好处，也可以不请我，到时候我便跟你说我请了你，你就欠了我一个人情;若是小兄弟答应我，日后我自然会等你。\\书生：啊，我明白了，这样就不会让你和我喝酒了。\\飞云：对对，小兄弟是有诚信的人，不如这样，你叫在下这样做，我也是这样做，我们就到此为止，日后互不相欠。\\书生：这...难道小兄弟不明白我的意思？\\飞云：我当然明白。\\书生：那我就明白了。\\飞云：怎么？难道你还不明白我的意思吗？\\书生：小兄弟，你相信我了？\\飞云：嗯。\\书生：那你应该明白我的意思，我不是不给你报酬的人，只是我不希望你因为我而耽误了自己的前程，你明白我的意思吗？\\飞云：我当然明白你的意思。\\书生：嗯，这一点我又何需多言？其实刚才我不是也说过了吗？我是一个有诚意的人，我很感谢你能让我明白了这么多，不过小兄弟你既然已经明白了我的意思，就表示你愿意接受我的这份礼物，我说到做到，你看怎么样？\\飞云：哦，那太好了，以后我若有什么困难，你就可以来找我了。\\书生：嗯，小兄弟，说得好。既然你已经接受了我的这份礼物，我也没有什么好说的了，不过你也要答应我，以后在没有我的允许下，你一定要记住，除了我，千万不要在别人面前提起你的家人。\\飞云：嗯。\\书生：我走了。\\飞云：小兄弟慢走，来日再见。\end{CJK}} \\ \hline
\end{tabular}
\end{table}

\begin{table}[]
\centering
\caption{We find that our \MODEL\ model is very good at writing fiction in different styles. This could be due to the fact that our train data contains lots of e-books. Source: \begin{CJK}{UTF8}{gbsn}
《古剑奇谭》and《传奇机长》, respectively.
\end{CJK}}
\begin{tabular}{l|l} \hline
\multicolumn{2}{c}{Fiction Generation} \\ \hline
Prompt                   &  
            
            \parbox{0.8\textwidth}{%
        \begin{CJK}{UTF8}{gbsn}
火神祝融取榣山之木制琴三把，名为凰来、鸾来、凤来。祝融对三琴爱惜有加，尤喜凤来，时常弹奏。凤来化灵，具人之形态，且能说人语。祝融大喜，请地皇女娲用牵引命魂之术将琴灵化为人身，以父子情谊相待，称之为太子长琴。
        \end{CJK}

        }            \\ \hline
Generation               &  \parbox{0.8\textwidth}{\begin{CJK}{UTF8}{gbsn}琴灵自然待祝融如同自己骨肉，祝融得琴通灵，聪颖过人，但性情刚烈，不近人情。有一次，长琴召来蛟龙偷入禁地，祝融得知此事后即将长琴囚禁。长琴不甘被囚禁，在蛟龙的帮助下逃离禁地。长琴在女娲宫修炼时，太子长琴于女娲宫中每日听琴，忽见一只白狐狸从禁地出逃，只听狐人惊呼“长琴”。长琴立刻命令宫娥去追狐人。宫娥追至狐人被禁地，那里地势低洼，长琴以为狐人已死，不肯追击。当宫人跑去救狐人时，只见那狐人已化为白凤，而宫娥则化为凤凰。\end{CJK}} \\ \hline\hline

Prompt                   &   
            
            \parbox{0.8\textwidth}{%
        \begin{CJK}{UTF8}{gbsn}
“话说...这个学员是什么来路，劳驾咱们的大队长亲自跟我交代啊？” 何常远一边抿着茶水，一边好奇地问曹进文。
曹大队长有些不耐烦：“都快六十，没两年退休的人了，哪来这么大好奇心！”
        \end{CJK}

        }            \\ \hline
Generation               &  \parbox{0.8\textwidth}{\begin{CJK}{UTF8}{gbsn}“嘿嘿...可能是吧。”曹大队长尴尬地干笑了两声,“我在这待了一辈子,就没见过这么不知趣的学员,再这么下去我的威信就要完了!”
“我看也未必。”\\
柳亦帆笑着说了一句,又低声对何常远道:“不要跟他硬碰,容易吃亏。”\\
“放心吧,他根本不可能在咱们大队长面前赢过我,就是跟他拼个两败俱伤也不可能,关键是要看那小子有什么本事。”\\
何常远一边整理着茶杯里的茶水,一边微笑着说道。\\
曹进文无奈地摆了摆手,把耳朵贴在门上,等待着院子里的动静。\\
就在这时,从院子里传来一声清亮的喊声:“蔡文高,我在这里!”\\
曹进文抬头看去,只见一名学员抱着一摞书从院子里走了出来。\\
他愣了愣,弯腰捡起那摞书,背在身上,大步走向了那张桌子。\\
待他走到近前,曹进文才看清桌子上的人,竟是一件崭新的制服,胸前的徽章不知道在什么时候被撸下来了。\\
“嘿嘿...还是大队长厉害啊,这制服真漂亮!”\\
曹进文冷笑了一声,“算你识相,我告诉你,咱们大队长可不是什么好东西,我看他跟你一样,都是来蹭饭的。”\\
“多谢大队长!”学员将手里的书本放下,又把制服展开,小心翼翼地捧到了桌子上。\\
曹进文将所有的注意力都集中到了学员身上,根本没注意到旁边那名学员已经被曹大队长刻意打了一个眼色,悄悄离开了院子。\\
看见学员完成这一套动作,曹进文满意地笑了笑,继续低头喝茶。\\
还没等学员将制服整理好,何常远忽然冒了出来:“队长!蔡文高这人好像不是我们辖区的,他好像是枫桥区那边来的。”\\
曹进文皱起了眉头:“啥?枫桥区?\end{CJK}} \\ \hline

\end{tabular}
\end{table}

\section{Conclusion}\label{sec:conclusion}

We have pretrained large-scale Chinese autoregressive language models named \MODEL, with up to 200 billion parameters. \MODEL\ has been developed under the MindSpore framework and trained on a cluster of 2048 Ascend AI processors. We believe there are many open problems in the field of large-scale PLMs:
\begin{itemize}
    \item Large-scale language models have demonstrated its promising few-shot capabilities in NLP tasks. However, the behaviors of such models are not systematically studied yet. How to make proper use of large PLMs and how to develop efficient few-shot algorithms remain open questions.
    \item Though effective, the computational cost for the inference of super large language models is still expensive. Thus it is worthwhile studying how to save the cost for the inference of large PLMs without sacrificing much of their performance. Model compression and acceleration of large PLMs could be an interesting topic.
    \item Training a even larger PLM with trillions of parameters will certainly bring more challenges to the both software and hardware sides. In addition, more efficient model structures such MoE~\cite{shazeer2017outrageously} or Switch Transformers~\cite{fedus2021switch} are also expected for relieving the computational cost of model training and inference.
    \item Pretrained multi-modal models integrating language, vision and speech data have attracted much attention recently~\cite{radford2021learning,lin2021m6}. Similar to the scaling law of language models, the performance of pertrained multi-modal models may also improve when the model sizes increase and more training data are collected. This is definitely a promising direction to explore. 
\end{itemize}

\section{Acknowledgements}
We thank Hanyang Wan, Qian Zhao, Yong Li, Zhou Cao, Yongqiang Lai, Zhijian Guo, Yue Wang, Zherui Chang, Junqiu Wei, Pingyi Zhou, Yulong Ao, Wenzhi Liu for their great support to this work. Also thanks for the support by the School of Electronics Engineering and Computer Science at Peking University, Central Software Institute and Noah's Ark Lab at Huawei Technologies, and Peng Cheng Laboratory. 

\clearpage

\bibliographystyle{unsrt}  
\bibliography{references}  

\end{document}